\documentstyle[11pt, lingmacros, fleqn, fullname]{article}

\newcommand{\titleinfo}{
{\begin{center} {\LARGE \bf
		Ellipsis and Higher-Order Unification		} \\[4ex]
{\bf 			   Mary Dalrymple			}\\
			     Xerox PARC				\\
		      Palo Alto, CA 94304, USA			\\
				and				\\
	  Center for the Study of Language and Information	\\
		        Stanford University			\\[2ex]
{\bf 			 Stuart M. Shieber			}\\
		    Division of Applied Sciences		\\
			 Harvard University			\\
		      Cambridge, MA 02138, USA			\\[2ex]
{\bf		       Fernando C. N. Pereira			}\\
		      AT\&T Bell Laboratories			\\
		     Murray Hill, NJ 07974, USA
\end{center}}}

\input{lfgmacros}

\newcommand{\lamabs}[2]{\lambda{#1}.\,{#2}}
\newcommand{\eqpunc}[1]{\makebox[0pt][l]{\qquad\rm{#1}}}
\newcommand{\fun}[1]{\mbox{\it #1\/}}
\newcommand{\prim}[1]{\underline{\mbox{\it #1}}}
\newcommand{\seq}[1]{\langle{#1}\rangle}
\newcommand{\hfilbreak}{\hfil\par\penalty10000\vspace{.5ex}\par\penalty10000}
\newcommand{\gloss}[2]{\leavevmode
        \vtop {
               \hbox {#1}
               \hbox {\deepstrut #2}}}
\newcommand{\deepstrut}{\hbox {\vrule height 2ex depth 1.2ex width 0pt}}

\newcommand{\wifeof}{\fun{wife-of}}
\newcommand{\husbandof}{\fun{husband-of}}
\newcommand{\paperof}{\fun{paper-of}}
\newcommand{\findingnude}{\fun{finding-nude}}
\newcommand{\refuse}{\fun{refuse}}

\newcommand{\golf}{\fun{golf}}
\newcommand{\before}{\fun{before}}

\begin{document}


\begin{titlepage}

\titleinfo

\thispagestyle{empty}

\vspace{.5in}

\begin{abstract}

We present a new method for characterizing the interpretive
possibilities generated by elliptical constructions in natural
language.  Unlike previous analyses, which postulate ambiguity of
interpretation or derivation in the full clause source of the
ellipsis, our analysis requires no such hidden ambiguity.  Further,
the analysis follows relatively directly from an abstract statement of
the ellipsis interpretation problem.  It predicts correctly a wide
range of interactions between ellipsis and other semantic phenomena
such as quantifier scope and bound anaphora.  Finally, although the
analysis itself is stated nonprocedurally, it admits of a direct
computational method for generating interpretations.

\end{abstract}

\vfill

\begin{center}
This article is available through the Computation and Language E-Print
Archive as cmp-lg/9503008, and also appears in {\em Linguistics and
Philosophy} 14(4):399--452.
\end{center}

\end{titlepage}


\tableofcontents

\newpage


\setlength{\parskip}{1ex plus 1ex minus 0ex}

\section{Introduction}

In this paper, we present a new method for characterizing the
interpretive possibilities generated by elliptical constructions in
natural language.  Unlike previous analyses, which postulate ambiguity
of interpretation or derivation in the full clause source of the
ellipsis, our analysis requires no such hidden ambiguity.  For
example, the ambiguity typically characterized as enabling ``strict''
versus ``sloppy'' readings of elliptical constructions does not arise
from a corresponding ambiguity as to whether the pronoun in the
antecedent clause is given a strict or sloppy interpretation; instead,
the ambiguity follows from the process of interpreting the elided
phrase on the basis of its unambiguous antecedent.  Further, the
analysis follows relatively directly from an abstract statement of the
ellipsis interpretation problem and applies to the interpretation of a
wide variety of elliptical constructions, including VP ellipsis, ``do
so'' and ``do it'' anaphora, gapping, stripping, and related
constructions involving recovery of implicit relations such as
``only'' modification and cleft constructions.  It predicts correctly
a wide range of interactions between ellipsis and other semantic
phenomena such as quantifier scope and bound anaphora.  Finally,
although the analysis itself is stated nonprocedurally, it admits of a
direct computational method for generating interpretations.

The analysis we present is intended to characterize the semantics of
constructions involving ellipsis.  Many interesting issues arise
regarding the syntax of ellipsis, and we will touch on some of these
issues; our main goal, though, is to characterize a method for
ellipsis interpretation.

\section{Basics}

\subsection{An Abstract Statement of the Ellipsis Problem}

We can provide an abstract and reasonably theory-neutral
characterization of ellipsis phenomena and their interpretation as
follows.  An elliptical construction involves two phrases (usually
clauses) that are parallel in structure in some sense.  The antecedent
or {\it source} clause is complete, whereas the {\it target} clause is
missing (or contains only vestiges of) material found
overtly in the source.  As a concrete example, which we will use as
our primary source of data in the paper, consider the verb phrase (VP)
ellipsis phenomenon, as in (1).

\enumsentence{ \label{first-eg}
Dan likes golf, and George does too.
}

\noindent The sentence is interpreted as meaning that Dan and George
both like golf.  The source clause, `Dan likes golf', parallels the
target `George does too', with the subjects `Dan' and `George' being
parallel elements, and the VP of the target sentence being vestigially
represented by the target phrase `does too'.\footnote{We emphasize
that although the bulk of the examples in this paper involve VP
ellipsis, the techniques can be applied to the semantic interpretation
of many other elliptical constructions.  This is in part because we do
not restrict the notion of parallel elements to the subjects of the
source and target clauses (see
Section~\ref{sec:nonconstituent-abstractions}).  Indeed, the
parallelism between the clauses need not even be purely syntactic (see
Section~\ref{sec:sem-parallel}).  The generality of this ellipsis
resolution method is one of its primary advantages.}

Given this abstract view of ellipsis, the problem of ellipsis
interpretation is just to recover a property of (or relation over) the
parallel element (respectively, elements) in the target that the missing or
vestigial material stands proxy for.  Of course, this property is not
arbitrary.  We know that the application of the property or relation
to the parallel elements in the source constitutes the interpretation
of the source clause.  In example~(\ref{first-eg}) above, we know then
that the property $P$ being predicated of George in the second
sentence is such that when it is predicated of Dan, it means that Dan
likes golf.  We might state this equationally as follows:

\[P(dan) = like(dan,\fun{golf})\]

\noindent A possible value for $P$ in this equation is the
property represented by the lambda term $\lamabs{x}{like(x,
\fun{golf})}$.  Predicating this property of George, we have
$[\lamabs{x}{like(x, \fun{golf})}](george)$ which reduces to
$like(george,\fun{golf})$.

In general, then, the abstract problem of ellipsis can be stated as
the problem of recovering solutions to the equation

\enumsentence{\label{general}
$P(s_1, s_2, \ldots, s_n) = s$
}
\noindent where $s_1$ through $s_n$ are the interpretations of the parallel
elements of the source, and $s$ is the interpretation of the source
itself.  (The determination of the parallelism itself is a separate
problem, about which more later.)  Once $P$ is determined,
$P(t_1,~t_2,~\ldots,~t_n)$ serves as the interpretation of the target,
where $t_1$ through $t_n$ are the interpretations of the corresponding
parallel elements of the target.

Not only is this an abstract characterization of the ellipsis problem,
it is essentially the entire analysis proposed in this paper.  It
constitutes an analysis because the equational statement of the
problem, together with some reasonable assumptions, determines
rigorously the sets of interpretations for target clauses, which
interpretations, we will see, correspond to the actual possible
interpretations of the target.

\subsection{Previous Analyses of Ellipsis}

It is important that ellipsis analyses (including the equational one
outlined above) allow for ambiguity in the target clause, that is, for
a set of relations to be made available by the source clause.  The
availability of multiple relations is attested in various phenomena in
which the target clause has multiple readings; it can be seen most
clearly in the distinction between strict and sloppy readings.  In the
sentence

\enumsentence{\label{eg:strict-sloppy}
Dan likes his wife, and George does too.
}

\noindent (under the reading in which the pronoun `his' refers to Dan)
the property predicated of George might be the property of liking
Dan's wife or the property of liking one's own wife.  In lambda
notation, these two properties are given by

\[\lamabs{x}{likes(x,~{\wifeof}(dan))}\]

\noindent and

\[\lamabs{x}{likes(x,~{\wifeof}(x))}\eqpunc{,}\]

\noindent corresponding to the strict and sloppy readings of the
sentence, respectively.\footnote{\label{fn:symbols}For brevity, we
represent the semantic interpretation of `Dan's wife' in this case as
$\wifeof(dan)$; the important thing to note about this representation
is that the semantics of the pronoun `his' is identical to that of its
antecedent, `Dan'.  Any other notation that has this property would do
as well here.  In particular, a treatment of possessives as
introducing bound variables along the lines of the quantifier
assumptions described in
Section~\ref{sec:interpretation-of-quantification} is possible and
perhaps preferable.  Notations such as $\wifeof(dan)$ can be thought
of as abbreviatory of such analyses.  We only use such notations for
cases in which the space of readings generated is unaffected.} As we
will see, the possibility of several available properties arises in
other cases as well.

Most previous analyses of ellipsis have allowed for the possibility of
multiple available properties by arranging for the source clause to be
ambiguous as to what property it makes available.  That is, in an
individual instance, the source clause is interpreted in one of
several possible ways, leading to the use of a particular property in
the interpretation; the property used in the target clause is
identical to the corresponding property in the source clause.  Dahl
\shortcite{Dahl:SI} was the first to draw a distinction between
approaches that place the ambiguity in the source clause (which he
called {\em strict identity} approaches) and those that place the
ambiguity in the process of recovering a property or relation for the
target (which he called {\em sloppy} or {\em non-strict identity}
approaches).  So as not to confuse the terminology here with that of
strict and sloppy readings, we will call the former kind of analysis
an {\it identity-of-relations} analysis, as opposed to a non-identity
analysis.

Under an identity-of-relations analysis, ambiguity of interpretation
in a target clause comes about because the source clause is ambiguous.
However, only a single relation is available from the source clause in
any given instance.  The relation that the source clause makes
available is, in most previous work, that associated with its VP,
though we emphasize that this is not a necessary condition for an
identity-of-relations analysis, nor do we believe it is a tenable
stance.

The multiplicity of interpretations for the elided phrase in an
identity-of-relations analysis may arise in various ways.  Purely
interpretive analyses \cite{GP:Anaph,Roberts:PhD} allow for multiple
semantic interpretations arising from an unambiguous syntactic
analysis of the source clause.  Partially interpretive analyses
involve either copying the syntactic tree from the source clause to
the target but requiring identical semantic interpretations for the
two VPs \cite{Williams:DLF}, or deleting the phrase structure of the
second tree under the constraint of identical interpretation
\cite{Sag:PhD}.  Finally, syntactic analyses may also allow the
semantic ambiguity to arise from underspecification in the copied
constituent \cite{Hellan:Anaph}.

The solutions therefore form a continuum, the ambiguity arising at more
or less superficial levels.  All of the analyses, however, share a
reliance on semantic ambiguity in the source clause.

Our solution to the question of what properties are made available can
be seen as lying at the far end of this continuum.  We eschew not only
syntactic ambiguity in the source clause, but semantic ambiguity
arising from any source, as a generator of the multiple readings of
elliptical constructions.  Instead, multiple solutions come about as a
natural result of directly stating the definition of the relation to
be applied in the target clause.

\subsection{The New Analysis and an Example}

As described earlier, the problem of extracting a relation from the
source clause can be stated equationally as

\[P(s_1, s_2, \ldots, s_n) = s\eqpunc{.}\]

\noindent In cases of VP ellipsis where the subjects of the source
and target are parallel, the equation is simply

\[P(s_1) = s\]

\noindent where $s_1$ is the interpretation of the subject of the source
clause.  By solving this equation for the unknown, $P$, we generate
the relation (or relations, if multiple solutions exist) that the
resolution of the ellipsis requires.

Huet's {\em higher-order unification} algorithm \cite{Huet:HOU}
provides a means of completely enumerating representations of the
solutions of such equations, under assumptions whose detailed
discussion we defer to Section~\ref{semantic-basics}.\footnote{The use
of higher-order unification to resolve elliptical constructions has
been independently noted by other researchers.  Pulman and Milward
implemented a prototype system that handled simple cases of VP
ellipsis and gapping along these lines \cite{Pulman:CLE}.  Pareschi
and Steedman's ``left conjunct decomposition'' operation, which is
used for the parsing of gapping constructions, bears a certain
resemblance to higher-order unification as well
\cite{Steedman:Gapping}.}

As an example of the use of equations to state a problem in ellipsis
resolution, consider (\ref{first-eg}) above, repeated here:

\enumsentence{
Dan likes golf, and George does too.
}

\noindent Recall that `Dan' and `George' are the parallel elements in
this example, and the semantic interpretation of `Dan likes golf' is

\enumsentence{
$like(\prim{dan},~\fun{golf})$ }

\noindent We have underlined what we will call {\em primary
occurrences} of the parallel element's interpretation, for reasons to be
clarified later.  In this case, the single occurrence of {\em dan} is
primary.  Any other occurrences will be referred to as {\em
secondary}.

To form an interpretation for the second conjunct, we require a
property $P$ that, when applied to the interpretation of the subject
of the first conjunct, will yield the interpretation of the first
conjunct as a whole.  This property will serve to generate the
interpretation of the target clause.  It will be applied to the
interpretation of the parallel element, that of the subject `George',
in the second clause.

We can state this requirement directly with the following equation, an
instance of the more general equation~(\ref{general}):

\enumsentence{\label{first-match-prob}
$P(dan) = like(\prim{dan}, \fun{golf})$}

\noindent The latter term is the interpretation for the source sentence;
the equation requires $P$ to be a property that, when predicated of
the subject interpretation {\it dan}, yields the first term.

A solution for an equation can be represented by a substitution of
values for the free variables in the equation that makes both sides of
the equation identical. For example, the following two alternative
substitutions solve equation~(\ref{first-match-prob}):

\enumsentence{ \it
$P \mapsto \lamabs{x}{like(\prim{dan}, \fun{golf})}$

$P \mapsto \lamabs{x}{like(x, \fun{golf})} $
}

\noindent The first substitution will be disregarded because it leaves
a primary occurrence in the result.  This constraint requiring {\em
abstraction of primary occurrences} comes about because the parallel
element in the target clause must play the primary role in the meaning
of the target.  We will have more to say about this in
Section~\ref{sec:primary-constraint}, especially as regards the
distinction between primary and secondary occurrences.  Given this
constraint, the only remaining value for $P$ is the property
$\lamabs{x}{like(x,\fun{golf})}$.  We can now use this function as the
interpretation of the elided VP in (1); with $P(george)$ as the
interpretation of the target clause, this gives the following
semantics for the sentence as a whole:\footnote{At this point, we can
ignore the distinction between primary and secondary occurrences.}

\enumsentence{ $like(dan, \fun{golf}) \wedge like(george, \fun{golf}) $ }

In summary, our analysis of the abstract problem of ellipsis
resolution, that is, generating appropriate properties to be used in
interpreting the target clause, is to state the problem equationally
based on the parallel structure in the two clauses and to solve the
equation using higher-order unification (under the constraint
requiring abstraction of primary occurrences).  The properties that
are generated as solutions to the equation are predicated of the
parallel elements in the target clause to generate the target clause
interpretation.

\subsection{Strict and Sloppy Readings}

As seen in the preceding example, the equation stating an ellipsis
interpretation problem may have several alternative solutions, which
the higher-order unification algorithm will generate.  Specifically,
when there are multiple occurrences of some subterm in a term,
multiple alternative substitutions for the relation formed by
abstracting out that subterm will exist.\footnote{Since multiple
occurrences of the same proper name do not necessarily co-denote, we
will use different constants as the representations of the denotata of
such occurrences.  For instance, the interpretation of the sentence

\enumsentence[(a)]{Dan likes Dan's wife.}

\noindent should be $like(dan_1, \wifeof(dan_2))$ where {\it
dan$_1$} and {\it dan$_2$} are separate constants that only
contingently co-denote.  As a consequence, an ellipsis like

\enumsentence[(b)]{Dan likes Dan's wife, and Bill does too.}

\noindent has only a strict reading, which accords with conventional
wisdom.} Consider sentence~(\ref{eg:strict-sloppy}), repeated here:

\enumsentence{
Dan likes his wife, and George does too.  \label{dan-george-wifes}
}

\noindent Let us assume the following interpretation for the first
conjunct:

\enumsentence{
$likes(\prim{dan}, \wifeof(dan)) $
}

\noindent The first occurrence of {\em dan}, which arises directly
from the parallel element, the subject `Dan', is primary; the
occurrence arising from the pronoun, which is not a parallel element,
is secondary.  Solution of the equation
$P(dan)=likes(\prim{dan},~\wifeof(dan))$ by higher-order unification
yields four ways of forming a property by abstracting out the
semantics of the subject.  The possible values for $P$ are

\enumsentence{
$\lamabs{x}{likes(\prim{dan}, {\wifeof}(dan))}$

$\lamabs{x}{likes(\prim{dan}, {\wifeof}(x))} $

$\lamabs{x}{likes(x, {\wifeof}(dan))}$

$\lamabs{x}{likes(x, {\wifeof}(x))}$ \label{dan-george-wifes-solns}
}

\noindent Again, the first two solutions fail the constraint on
abstraction of primary occurrences.  Either of the other two remaining
properties yields a possible interpretation of the target clause.  The
first gives rise to what has been called the strict reading of the
second conjunct, while the second gives rise to the sloppy reading:

\enumsentence{
$\lamabs{x}{likes(x, {\wifeof}(dan))}(george) = likes(george, \wifeof(dan))$

(George likes Dan's wife.)

$\lamabs{x}{likes(x, {\wifeof}(x))}(george) = likes(george, \wifeof(george)) $

(George likes George's wife.)
}

\subsection{Constraints on Relation Formation}

\label{sec:primary-constraint}

As illustrated in the foregoing examples, we use the distinction
between primary and secondary occurrences of a term to constrain the
acceptable solutions of an ellipsis interpretation equation.  We
define a primary occurrence as an occurrence of a subexpression in the
semantic form directly associated with one of the parallel elements in
the source clause; we assume that the notion of ``directly
associated'' is sufficiently well-defined for the purposes at hand.
We then require that the solution process preserve the primacy of
occurrences, with the consequence that solutions must abstract over
all primary occurrences in the source. In other words: {\em Solutions
must not include primary occurrences.}  In
example~(\ref{dan-george-wifes}), this constraint removes from
consideration the first two putative solutions
in~(\ref{dan-george-wifes-solns}).

If the constraint were not in force, the following readings would be
produced for `... and George does too':

\enumsentence{
$\ast$ ... and Dan likes Dan's wife.

$\ast$ ... and Dan likes George's wife.
}

\noindent These are just the readings where the parallelism between
the clauses has been disregarded.  Thus, the constraint is a reflex of
the inherent parallelism in elliptical constructions.

The existence of this constraint means, not surprisingly, that it is
necessary to retain a connection between the syntactic and the
semantic representation of the source sentence.  By maintaining this
connection, we can ensure that the solutions produced by higher-order
unification satisfy the constraint that parallelism must be maintained
by abstracting out of parallel positions.\footnote{Solutions involving
vacuous abstraction, such as

\[\lamabs{x}{likes(\prim{dan},~{\wifeof}(dan))}\eqpunc{,}\]

\noindent are ruled out where
necessary as special cases of this more general constraint.
A direct prohibition against vacuous abstraction might be
too strong, since verb phrase ellipsis is possible even in cases where
the subject of the source clause is pleonastic and makes no
semantic contribution (examples due to Ivan Sag):

\eenumsentence{
\item John said it would rain, and it did.
\item John said there would be trouble, and there was.}

\noindent Suppose the interpretation of the former example (ignoring
tense and aspect as usual) were

\[said(john,rain)\eqpunc{.}\]

\noindent Then the
second-order matching problem induced by the ellipsis would be

\[P(\Delta) = rain\eqpunc{.}\]

\noindent (We use the symbol $\Delta$ for specifying the
interpretation of pleonastic elements, following \namecite[page
221]{GKPS:GPSG}.)  The requirement of abstraction of primary
occurrences would still allow a binding for $P$ involving vacuous
abstraction, namely $\lamabs{x}{rain}$.  The elliptical clause would
therefore be interpreted as $rain$.
}

\subsection{Constraints on Parallelism}

One of the distinguishing features of our analysis is that the
ellipsis resolution problem is separated into two subtasks: a prior
determination of the parallel structure of source and target, and
consequent formation of the implicit relation to be used in the
target.  We have been addressing the latter subtask primarily, and
will continue to do so, but we digress to mention some perhaps obvious
facts about the parallelism determination that might get lost in the
sequel.

 The task of determining the parallel structure of two
clauses is far more subtle, and less syntactic, than a cursory
examination exposes.  (We discuss this issue further in
Section~\ref{sec:sem-parallel}.)  For this reason, the division of the
ellipsis problem into two parts---separating parallelism determination
from relation formation---allows a simpler description of relation
formation and a more appropriate characterization of the problem of
determination of parallelism.  Nonetheless, this paper does not
provide a theory of parallelism; previous attempts have been far too
restrictive, limiting themselves to purely syntactic criteria.  The
wide range of possibilities for parallelism described in
Section~\ref{sec:sem-parallel} indicate that the process is not a
purely linguistic one.  As an extreme example, we describe in
Section~\ref{sec:sem-parallel} cases of parallelism with no linguistic
source whatsoever.

 Our emphasis in this paper on issues in relation
formation and our liberal view of parallelism determination should
not, of course, be taken to imply that no constraints apply to the
task of determining parallelism.  For example, parallelism must
respect stativeness of verbs (\ref{ex:parallel-constraints}a) and
pleonasticity of noun phrases (\ref{ex:parallel-constraints}b).

\eenumsentence{\label{ex:parallel-constraints}
\item $\ast$ Dan likes golf and George is too.
\item $\ast$ It is raining and George is too.
}

\noindent Depth of embedding imposes constraints as well.

\enumsentence{
$\ast$ The mayor of Washington left, and New York did too.}

\noindent (We ignore the nonsensical reading in which the city of New
York was the agent of the leaving action.)  Similarly, the sentence

\enumsentence{It is obvious that Dan is happy, and George is too.}

\noindent (pointed out to us by Mats Rooth) can only be interpreted as
meaning that George is happy, not that it is obvious that George is
happy.  If the obviousness is included, the parallelism would have to
hold between `George' and `it', not `Dan'.

Such constraints hold not only in VP ellipsis, but also in gapping,
stripping, comparative deletion and other elliptical constructions.
Thus, not all constraints on readings of elliptical sentences follow
from the relation formation issues that are the primary topic of this
paper.

Furthermore, there are syntactic constraints that apply differentially
to different ellipsis constructions.  Even among constructions eliding
VPs, such as the ``do'', ``do so'', and ``do too'' variants, syntactic
distinctions can be found.  For instance, as noted by \namecite[page
177]{Haik:PhD}, these variants differ in their
grammaticality in antecedent-contained-ellipsis contexts:

\eenumsentence{
\item  John greeted every person that Bill did.
\item * John greeted every person that Bill did so.
\item ? John greeted every person that Bill did too.
}

\noindent These fine syntactic distinctions are not addressed by the
present analysis, which attempts to make clear only the space of
semantic interpretations.

Of course, not all elements in the target clause must be analyzed as
parallel to some element in the source.  For instance, adverbial
phrases can be viewed as modifying the target directly.  This
possibility is exemplified by the following sentence:\footnote{We are
indebted to an anonymous reviewer for this example.}

\enumsentence{
Jim couldn't open the door, but Polly did with her blowtorch.}

\noindent No empty adverbial modifier need be posited in the source
clause; the instrumental modifies the target sentence directly.  (Such
elements can be made parallel in other cases, however; see
Section~\ref{polarity} for examples.)

\subsection{Formal Semantic Background}
\label{semantic-basics}

We outline here the formal machinery underlying the semantic
analyses used in this paper.

Meanings of phrases are to be represented by terms of a typed
higher-order system with lambda abstraction. Since we are not
concerned here with intensional phenomena, we will just need the basic
types $e$ (entities) and $t$ (truth values). Two type constructors
will be used: $\rightarrow$ to form the type $T\rightarrow T'$ of
functions mapping arguments of type $T$ to results of type $T'$, and
$\times$ to form the type $T\times T'$ of pairs $\seq{t,t'}$ such that $t$
has type $T$ and $t'$ has type $T'$.

We will use various elementary concepts from the lambda calculus,
specifically the notions of {\it free} and {\it bound} occurrences of
variables, and of {\it substitution} of a lambda term for a variable.  We
will notate the substitution of term $N$ for all free occurrences of
$x$ in $M$ by $M[x \mapsto N]$.  We require, as is typical, that
substitution rename bound variables in $M$ appropriately to avoid
capture of free variables in $N$.  The reader is urged to refer to
Hindley and Seldin \shortcite{HindleySeldin:Lambda} for precise definitions
of these notions.

We have proposed codifying the ellipsis interpretation problem using
expressions equating terms that represent phrase meanings. What counts
as a solution to such an equation depends crucially on what notion of
equality between terms we are considering, or, in other words, on when
we consider that two terms denote the same semantic object. One
salient notion of equality is that of $\alpha\beta\eta$ {\em
interconvertibility} \cite{HindleySeldin:Lambda}, which captures
formally the intuitive notion that two terms can represent the same
``recipe'' for calculating a function.  Specifically, two terms are
considered equal if one can be converted to the other by repeated
application of the following rules and their inverses:
\begin{description}
\item[$\alpha$ conversion:] convert $\lamabs{x}{M}$ to
$\lamabs{y}{(M[x \mapsto y])}$, that is, the names of bound variables
are immaterial to their meaning. The two terms are said to be {\em
alphabetic variants}.
\item[$\beta$ conversion:] convert $[\lamabs{x}{M}](N)$ to
$M[x\mapsto N]$. This represents formally the operation of applying a
function to an argument.
\item[$\eta$ conversion:]
convert $\lamabs{x}{M(x)}$ to $M$ when $x$ does not occur free in $M$.
\end{description}

\subsubsection[Interpretation of Ellipsis Resolution
Equations]{Interpretation of Ellipsis Resolution
Equations\footnotemark}\footnotetext{We are indebted to Mark Johnson
and an anonymous reviewer for crystallizing these issues in our minds
and for organizing the structure of possible responses.  The
particular statements made here are our own, however, and should not
be interpreted as indicative of their opinions.}

We have claimed that the meaning of

\enumsentence{\label{ex:golf}
Dan likes golf, and George does too.}

\noindent is

\enumsentence{$like(dan, \golf) \wedge P(george)$}

\noindent where the equation

\enumsentence{$P(dan) = like(dan, \golf)$}

\noindent must be satisfied.  It is not obvious that the equation in
\ex{0} is semantically interpretable, as opposed to being a recipe for
invoking a formal procedure, higher-order unification, with no
underlying meaning in and of itself.  Clearly, the combined meaning of
\ex{-1} and \ex{0} is not equivalent to

\[ \exists P. like(dan, \golf) \wedge P(george) \wedge P(dan) =
	   like(dan, \golf) \eqpunc{.}\]

\noindent This would merely require that $P$ be such that it is true
of Dan if Dan likes golf.  Since the first conjunct states that Dan
does in fact like golf, $P$ need only be a true property of Dan.  The
entire formula then requires only that George possess some property,
any property, that Dan possesses, which would give an incorrect
interpretation for the target sentence.  The equation is not to be
interpreted, then, as codenotation in a model.

Instead, we want ellipsis to be more content-independent, in that the
property should be such that the equation holds whether or not Dan
happens to like golf.  It should be independent of the particulars of
a given model, that is, it should hold in any model in a suitable
class of models.  But we have to be careful about the choice of model
class.  Even necessary truths should not codenote over the class of
models in which the ellipsis equation is interpreted.  Otherwise, the
sentence

\enumsentence{
Every square has four sides, and every rhombus does too.}

\noindent would be subject to an interpretation where what is
predicated of every rhombus is some property of squares that is true
of them in whatever models assign squares four sides.  But these, {\it
ex hypothesi}, include all the models, so any property true of all
squares (such as having four equal angles) would do.  The sentence
might mean, then, that every rhombus has four equal angles.  Of
course, it does not.  Similarly, logical tautologies must not be
valid; sentence~(\ref{ex:golf}) does not, for example, mean that
George likes golf and either it is raining or it is not raining, even
though this is logically equivalent to George liking golf.

The class of models, then, in which the ellipsis equation is held as
valid is very weak.  In fact, for higher-order unification to be an
appropriate procedure for determining valid instances, the valid
equations must be exactly those whose two sides are $\alpha \beta
\eta$-interconvertible.  Higher-order unification finds the most
general substitutions of terms for the free variables in an equation
that make the equation valid.

\namecite{Friedman:Equality} demonstrates that such equalities are
exactly those that hold in all extensional models for the typed lambda
calculus (without interpreted constants), and also exactly those that
hold in any model consisting of all the higher-order functions over
some infinite base set, with application interpreted as function
application.  Although the logic that we presuppose is augmented with
a full first-order quantificational logic (and presumably, for
intensional phenomena, would be an intensional logic), recall that the
tautologies of this logic are not required to hold for the purposes of
interpreting the ellipsis equations; the symbols of the logic
($\forall$, $\wedge$, etc.) can be viewed as uninterpreted function
symbols of the appropriate type.  The only structure that the model
manifests is, then, the structure arising from the categorial or type
structure of the language, together with the reasonable requirement of
extensionality.

Thus, the semantic invariants in ellipsis resolution are those that
follow from the type structure---the function-argument
relationships---of natural language, and not from any contingent or
even necessary truths.  This accords with intuition, in that the
felicity of ellipsis does not depend on the meaning of the words in
the source sentence (though the elided property does), but does depend
on their type structure.  An ellipsis equation is not merely a recipe
for a syntactic process.  It has a meaning, but the meaning must be
taken in a different, and much more profligate, model than that of the
interpretation of the sentence itself.  The ellipsis equation reflects
semantic facts about the sentence, but just at the gross level of
function-argument structure.

There is one remaining problem, however, for this view of ellipsis
equations---the issue of primary occurrences.  The primary occurrence
notation serves to couple the ellipsis equations to the choice of
parallel elements, and provides a way of forcing abstraction over the
meanings of the parallel elements.  Intuitively, the distinction
between primary and secondary occurrences is clear: a primary
occurrence corresponds to a distinguished semantic role in the
situations described by the source and target clauses.  At present,
however, we have no way of making precise the intuitions that led to
the primary occurrence notation; that is, we know of no semantic
correlate to the equational system with primary occurrence notation
under the related constraint on abstraction.  It remains for future
work to reconstruct the semantical foundations of this variant of
higher-order unification.  Although we have some ideas as to how such
a reconstruction might proceed, it is premature to discuss them here.

\subsubsection{Higher-order unification}

The unification problem for terms in a logical system is the problem
of finding substitutions for the free variables of two terms $t$ and
$t'$ that make the terms equal.  Such a substitution is called a {\em
unifier} of $t$ and $t'$. A unifier $\sigma$ is {\em more general}
than another unifier $\sigma'$ if there is a nontrivial substitution
$\tau$ such that $(t \sigma) \tau=t \sigma'$, where $t \sigma$ is
the result of applying substitution $\sigma$ to term $t$. Informally,
a more general unifier will leave more free variables, or make fewer
variable identifications, than a less general one. Unifiers that are
most general represent the solutions of a unification problem in their
simplest form, since any less general unifier (solution) can be
obtained from the output of a most general one by additional
substitutions of terms for free variables. As is well known, the
unification problem for first-order terms, and the related unification
problem for certain kinds of feature graphs, admit of unique most
general unifiers (up to variable renaming). However, this is not the
case for higher-order unification, in which variables can range over
functions of arbitrary order rather than just over [first-order]
individuals. This multiplicity of unifiers corresponds to the
possibility of multiple alternative interpretations for elided
material.

Huet's higher-order unification algorithm enumerates the unifiers of
higher-order terms in a typed $\lambda$-calculus of the kind we are
using. Because higher-order unification is in general undecidable,
given two terms, the algorithm will either stop without finding any
unifiers (the terms are not unifiable), generate successive [most
general] unifiers (possibly without end), or run forever without
producing any unifiers.  This computational property of higher-order
unification has not been problematic on the cases we have examined
that are engendered by ellipsis resolution, however, and there are
several reasons why this might be so.

First, many of the equations arising in ellipsis resolution fall under
the subcase called {\it second-order matching}.  In the second-order
subcase of unification, variables range only over individuals and
first-order functions.  The simpler matching problem occurs when the
substitution need be applied only to one of the terms, that is, $t_1
\sigma = t_2$.  Huet and Lang \shortcite{HuetLang:SecondOrder} use
second-order matching as a way of applying program transformations in
a manner reminiscent of the method used for ellipsis interpretation in
this paper.  Second-order matching is, fortunately, decidable, and
Huet and Lang provide an algorithm, which is an adaptation of Huet's
more general algorithm for the subcase of interest.

Furthermore, many of the equations we will be interested in solving
are of the schematic form

\[P(s_1, s_2, \ldots, s_n) = s_0\]

\noindent where the $s_i$ are all ground, that is, contain no free
variables.  The special case of second-order matching engendered by
instances of this schema is computationally even more tractable.
There are only a finite number of unifiers and these can be simply
constructed as follows: Construct a term $s$ from $s_0$ by replacing
zero or more instances of the $s_i$ by $x_i$.  For each such $s$,
construct a possible binding for $P$ given by $\lamabs{x_1}{\ \cdots
\lamabs{x_n}{s}}$.  Clearly, there are at most $2^c$ such unifiers
where $c$ is the number of occurrences of the $s_i$ in $s_0$, and
these can be enumerated efficiently (in time linear in the output
length).

Finally, although certain phenomena require use of the more general
higher-order unification (as the examples in
Section~\ref{sec:type-raising}), the bulk of the cases considered in
this paper rely on the ground subcase of second-order matching, and
are therefore less computationally problematic.  Of course, even the
higher-order cases we consider may turn out to fall into a
computationally reasonable subclass; further inquiry in this area
would be useful.

\subsubsection{Interpretation of quantification and long-distance
dependencies}
\label{sec:interpretation-of-quantification}

Before characterizing the interaction between our analysis of ellipsis
and various other semantic phenomena, we must first lay out an
approach to semantic interpretation---quantifier scoping in
particular---in which to couch the discussion.  For the most part, the
particulars of the method for characterizing quantifier scoping are
relatively unimportant; the analysis could be stated in terms of
Cooper storage, say, or even quantifier raising.\footnote{The
incorporation of an ellipsis analysis such as ours into a
transformational framework is quite conceivable, merely requiring the
ability to form abstractions over the syntactic objects representing
semantic construals of sentences, that is, LF trees.  The intrinsic
portion of the analysis is its use of an equational framework for
declaratively characterizing ellipsis resolution, not its use of
particular logics for the representation of meanings.  Nonetheless,
the use of typed lambda calculus allows us to directly state our
analysis with a minimum of extraneous machinery.} We will use here a
variation of a method for interpreting quantifier scoping and
long-distance dependencies developed by Pereira
\shortcite{Pereira:SemComp}.  For those readers unfamiliar with this
method, we provide some examples later in this section.

In general, the interpretation of a phrase will have the form
$\Gamma\vdash m$ where $\Gamma$ is a set of {\em assumptions}
analogous to a quantifier store in the Cooper storage method
\cite{Cooper:Quant} and $m$ is a matrix term in which free variables
introduced by the assumptions in $\Gamma$ may occur.

The assumptions used for quantifier scoping are triples of the form
$\langle q\;x\;p\rangle$ where $q$ is a determiner meaning, $x$ is a
free variable, and $p$ is a term of type $t$ in which $x$ is
free.\footnote{Formally similar store elements have been used in
quantifier scoping systems such as those of Schubert and Pelletier
\shortcite{SchubertPelletier82} and Hobbs and Shieber
\shortcite{HS:Quant}.} The assumption $\langle q\;x\;p\rangle$ is said
to {\em introduce} variable $x$.  A quantified noun phrase is
interpreted as a variable introduced by an assumption whose first
component is the meaning of the noun phrase's determiner and whose third
component represents the meaning of the noun phrase's nominal.

Instead of the usual generalized quantifier type $(e\rightarrow
t)\rightarrow(e\rightarrow t)\rightarrow t$ for determiner meanings,
we will use the {\em pair quantifier} type $(e \rightarrow t\times
t)\rightarrow t$. In other words, the meaning of a determiner takes a
function from entities to pairs of truth values and yields a
truth value. For example, the meaning of {\em every} will be the
function that assigns `true' just to those functions that take each
entity to a pair whose second component is true whenever its first
component is.

Our somewhat unusual type for determiner meanings, which is needed in
our analysis of antecedent-contained ellipsis
(Section~\ref{antecedent-contained}) can be understood as a lambda
calculus implementation of some aspects of Discourse Representation
Theory (DRT) \cite{Kamp:TSR,Heim:PhD}.\footnote{Dynamic Montague
grammar \cite{GroenendijkStockhof:Dynamic,Chierchia:Dynamic} provides
a more complex and possibly more general approach to incorporating
some of the aspects of DRT into a compositional framework.}
Specifically, the interpretation of

\enumsentence{John greeted every person.\label{ex:greet-every-person}}

\noindent will be for us

\enumsentence{$every(\lamabs{x}{\seq{person(x),greet(j,x)}})$
	\label{pair-quant}}

\noindent which we will abbreviate as
\[every(x,person(x),greet(j,x))\eqpunc{.}\]
\noindent This interpretation can be directly related to the discourse
representation structure (DRS) for the same sentence:
\enumsentence{\evnup{
\fbox{\fbox{\parbox{.7in}{
              \ \ \ \ \ \it x\\
              person(x)}} $\Rightarrow$
\fbox{\parbox{.65in}{\it greet(j,x)}}}}}
The discourse marker $x$ corresponds to the variable to be abstracted,
the left inner DRS to the first element of the pair in
(\ref{pair-quant}) (the quantifier restriction), the right inner DRS to
the second element of the pair (the quantifier scope) and the arrow to
the determiner meaning $every$. The referential connection established
by the discourse marker $x$ in the DRS is simulated in our analysis by
the simultaneous abstraction of the variable $x$ over both the
restriction and the scope of the quantifier.

It is straightforward to show that there is a one-to-one
correspondence between Barwise-Cooper generalized quantifiers and pair
quantifiers.  Indeed, such a correspondence is established by two
functionals, $\cal G$, mapping pair quantifiers to generalized
quantifiers, and $\cal P$, mapping generalized quantifiers to pair
quantifiers, satisfying ${\cal G}({\cal P}(Q)) = Q$ and ${\cal
P}({\cal G}(P))=P$.  $\cal G$ and $\cal P$ are defined as:
\begin{eqnarray*}
{\cal G}(P) & = & \lamabs{r}{\lamabs{s}{P(\lamabs{x}{\seq{r(x),s(x)}})}} \\
{\cal P}(Q) & = &
\lamabs{p}{Q(\lamabs{u}{\fun{fst}\/(p(u)),\lamabs{v}snd(p(v)))}}
\end{eqnarray*}
where $\fun{fst}\/(\seq{x,y})=x$ and $snd(\seq{x,y})=y$.

For a derivation to be considered complete, all assumptions must be
{\em discharged}. We will exemplify this process with
sentence~(\ref{ex:greet-every-person}).

The quantified noun phrase `every person' is given the interpretation

\[\langle every\;x\;person(x)\rangle \vdash x\eqpunc{.}\]

\noindent that is, the meaning of the noun phrase is $x$ under the
assumption to the left of the $\vdash$.

The verb meaning $\lamabs{o}{\lamabs{s}{greet(s,o)}}$ applied to the
NP meaning yields a VP meaning, still under the above
assumption:

\[\langle every\;x\;person(x)\rangle \vdash \lamabs{s}{greet(s,x)}\eqpunc{.}\]

\noindent Application of this VP meaning to the subject meaning
$\vdash~j$ results in this sentence meaning:

\[\langle every\;x\;person(x)\rangle \vdash greet(j,x)\eqpunc{.}\]

Discharging the quantifier assumption involves applying the quantifier
$every$ to the result $\lamabs{x}{\langle(person(x),
greet(j,x))\rangle}$ of abstracting $x$ over the pair consisting of
the restriction and the scope of the quantifier. The resulting
interpretation is \[\vdash every(x,person(x),greet(j,x))\] that is, a
sentence meaning free of undischarged assumptions. When several
quantifier assumptions are introduced, there is the option of
discharging them in several different orders, leading to alternative
quantifier scopings for the sentence.

In Pereira's original system, the treatment of quantifier assumptions
is semantically justified by showing how such derivations are
convenient shorthand for derivations in the Curry system of semantic
combination containing only the operations of functional application
and abstraction. In the present variant, we could carry out a similar
justification in a system that would include pairing in addition to
functional application and composition
\cite{Lambek80,Vanbenthem:Categorial}.

The treatment of the semantics of long-distance dependencies is
handled similarly by introducing and discharging assumptions.
Again, the form of these introduction and discharge rules could be
justified on the basis of functional application and abstraction.  As
an example of a derivation involving a long-distance dependency, we
consider the example

\enumsentence{John greeted every person that arrived.}

\noindent The trace in the subject position of the relative clause can
be thought of as introducing a $bind$ assumption for a new variable

\[ \langle bind\; x \rangle \vdash x \]

\noindent which serves as the argument to $\lamabs{s}{arrive(s)}$ (the
interpretation of `arrived'):

\[ \langle bind\; x \rangle \vdash arrive(x) \eqpunc{.} \]

\noindent The relative clause being completed, we can now discharge
the bind assumption.  We do so by forming a higher-order predicate by
abstracting the matrix, conjoined with a place-holder for the modified
nominal, and abstracted by the bound variable.

\[ \vdash \lamabs{N}{\lamabs{x}{N(x) \wedge arrive(x)}} \]

\noindent This relative clause meaning serves as a function over the
nominal meaning $\lamabs{y}{person(y)}$.

\[ \vdash \lamabs{x}{person(x) \wedge arrive(x)} \]

\noindent Finally, the rule for combining a quantifier $every$ with a
predicate forms a quantifier assumption over a new variable.

\[ \langle every\; z\; person(z) \wedge arrive(z) \rangle \vdash z \]

{}From here, the derivation continues as before, yielding the sentence
meaning (before the quantifier is discharged)

\[ \langle every\; z\; person(z) \wedge arrive(z) \rangle \vdash greet(j,z) \]

\noindent and the scoped meaning is

\[ \vdash every(z, person(z) \wedge arrive(z), greet(j, z)) \eqpunc{.}\]

This completes the background information on the formal semantics we
will presume in the remainder of the paper.

\section{Interactions with Other Phenomena}
\label{sec:interactions}

The approach to ellipsis resolution that is advocated here displays
differences from previous approaches in its handling of various
phenomena.  We will discuss how our analysis differs from
identity-of-relations analyses in general, and certain particular
instances thereof, by briefly examining the predictions of the
analyses with respect to the following phenomena:

\begin{description}

\item[Non-constituent abstractions:]  There are many cases in which
the relation constructed from the source clause does not correspond in
any straightforward fashion to the interpretation of some syntactic
constituent: for example, when it must take more than one argument.
For instance, the tense and aspect as well as the subject are
abstracted in the sentence
\begin{quote}
Dan is running for president, and George did last term.
\end{quote}
Examples demonstrate other nonstandard abstractions as well. Such
cases are especially problematic for identity-of-relations analyses in
which the relation  is necessarily associated with some
constituent such as the VP in the source clause.

\item[Multiple property extraction:] In some cases, a single sentence
serves as the antecedent for two subsequent instances of ellipsis
involving different parallel elements:
\begin{quote}
John finished reading the poem before Bill did, and the short story too.
\end{quote}
This sentence has a reading on which John finished reading both the
poem and the short story before Bill finished reading the poem.  This
is problematic for identity-of-relations analyses in which only a
single property is available in any given instance for the
interpretation of subsequent elided phrases.

\item[Cascaded ellipsis:]  Analyses differ as to what readings are
predicted for sentences containing multiple elliptical clauses in
which the interpretation of one elided constituent depends partially
or entirely on the interpretation of another elided constituent.  An
example is:
\begin{quote}
John realizes that he is a fool, but Bill does not, even though
his wife does.
\end{quote}

\item[Interaction with quantifier scoping:]  As is well known, the
ambiguities following from varying quantifier scope possibilities
interact with ellipsis resolution possibilities.  For instance, in the
sentence
\begin{quote}
John greeted every person when Bill did.
\end{quote}
two readings are possible, depending on whether the universal quantifier
has wide scope over both the main and subordinate clause, or
quantifies separately in each clause.  But in
\begin{quote}
John greeted every person that Bill did.
\end{quote}
only a wide scope reading is available.

\end{description}

We discuss each of these phenomena below, and demonstrate that our
approach constructs appropriate solutions.  In the succeeding section,
we discuss in detail an example sentence which illustrates differences
among a number of analyses of ellipsis that have been proposed in the
past.  Finally, we turn to problematic cases for this and other
analyses.

\subsection{Non-Constituent Abstractions}
\label{sec:nonconstituent-abstractions}

There are many instances in which the interpretation of elided phrases
does not correspond to the interpretation of a syntactic constituent
in the source clause.  The most obvious cases include the elliptical
constructions of gapping or stripping.  But VP ellipsis provides
examples as well.  For instance, there are cases in which a deeply
embedded constituent induces a sloppy reading; in other cases,
relations are formed with multiple parallel elements in the source and
target clause; in still other cases, as discussed in
Section~\ref{sec:sem-parallel} below, the parallelism between the
elements in the source and target clause is not syntactic, but
semantically or pragmatically induced.  These cases are problematic
for identity-of-relations approaches in general, since such approaches
would have to make available a very large number of different semantic
analyses for each source clause, some of them otherwise unmotivated,
to allow for all of the possible interpretations that might need to be
provided for subsequent ellipsis.  They are particularly problematic
for identity-of-relations analyses in which the interpretation
provided by the source clause corresponds to the translation of a
syntactic constituent in the source.

\subsubsection{Sloppy readings with embedded antecedents}

The primary argument given by Reinhart \shortcite{Reinhart:Anaph} for
the distinction between bound variable and referential pronouns is the
requirement that bound variable pronouns must be c-commanded by their
antecedents.  She uses this requirement to predict that the following
example has only a strict reading:

\enumsentence{\label{reinhart-eg}People from LA adore it and so do
people from NY.  [Reinhart's (17a), page 150]}

\noindent Reinhart proposes a requirement that a pronoun must be
c-commanded by its antecedent if the antecedent is a quantifier;
further, she claims that a pronoun giving rise to a sloppy
interpretation must be c-commanded by its antecedent.  For Reinhart,
then, the availability of a sloppy reading correlates with the
possibility of a bound-variable interpretation of a pronoun, and she
requires a c-command relation for this interpretation to be possible.
This restriction simplifies the task of an identity-of-relations
analysis because it reduces the number of cases in which a sloppy
reading is available.  An analysis postulating ambiguity of pronoun
interpretation for only this restricted set of cases seems
methodologically more plausible.

However, Reinhart herself \shortcite[page 178]{Reinhart:Anaph} notes certain
counterexamples to this correlation, cases where a sloppy reading is
available even when c-command does not hold:

\eenumsentence{
\item Felix$_i$'s mother thinks he$_i$'s a genius and so does
Siegfried's mother.  [Reinhart's (8a)]

\item We'll discuss Rosa$_i$'s problems with her$_i$ parents and
Sonya's problems too.  [Reinhart's (8b)]
}

\namecite{Wescoat:StrictSloppy} notes a number of more extreme cases of
sloppy readings involving non-c-commanding, embedded constituent
antecedents, such as:

\eenumsentence{\label{eg:wescoat1}
\item The policeman who arrested John failed to read him his rights, and so
did the one who arrested Bill.

\item The person who introduced Mary to John would not give her his phone
number, nor would the person who introduced Sue to
Bill.}

\noindent Wescoat claims, and we agree, that sloppy readings are
possible with these sentences; that is, that the following readings
are available---perhaps even preferred---for them:

\eenumsentence{
\item The policeman who arrested John failed to read John John's
rights, and the one who arrested Bill failed to read Bill Bill's rights.

\item The person who introduced Mary to John would not give Mary John's
phone number, and the person who introduced Sue to Bill would not give
Sue Bill's phone number.}

Hirschberg and Ward
\shortcite{HirschbergWard:StrictSloppy} have obtained experimental
evidence consistent with these examples that the c-command criterion
posited by Reinhart and counterexemplified by Wescoat is not a general
requirement for sloppy readings.  For example, their data show that as
many as one-third of a group of  test subjects preferred
the sloppy reading of
\enumsentence{People from Los Angeles think it's a scary place to
live, and so do people from New York.  [Hirschberg and Ward's (31)]}
which parallels sentence~(\ref{reinhart-eg}) closely.

On our analysis, barring any stipulated prohibition, there is no
obstacle to forming relations abstracted over arbitrarily deeply
embedded positions.  In skeletal form, the analysis of such an example
would proceed as follows.  Suppose the source clause of
(\ref{eg:wescoat1}b) is interpreted as\footnote{See
Footnote~\ref{fn:symbols} for a discussion of the status of the
function symbols $pwi$ and $phone$.}

\[\refuse(pwi(m,j), give(m, phone(j)))\]

\noindent where $pwi(x,y)$ is the person
who introduced $x$ to $y$, and $phone(x)$ is $x$'s phone number. The
parallel elements in the construction are, respectively, `the person
who introduced Mary to John' and `the person who introduced Sue to
Bill'; `Mary' and `Sue'; `John' and `Bill'.  Thus, the appropriate
equation to solve is

\[P(pwi(m,j),m,j) =
\refuse(\prim{pwi(\prim{m},\prim{j}}), give(m, phone(j)))\eqpunc{.}\]

\noindent The sloppy reading is engendered by the following unifying
substitution for $P$:

\enumsentence{\label{recovered-rel}
	\(\lamabs{x}{\lamabs{y}{\lamabs{z}{\refuse(x, give(y,
	phone(z)))}}}\eqpunc{,}\)}

\noindent which, when applied to the interpretations of the parallel
elements in the target, yields the target interpretation

\[P(pwi(s,b),s,b) =
\refuse(pwi(s,b),give(s,phone(b)))\eqpunc{.}\]

Note that the recovered relation~(\ref{recovered-rel}) is not a
relation corresponding to a conventional interpretation for the
VP---or any other constituent---in the source clause.  On an
identity-of-relations analysis, such relations would be available only
by virtue of their use in the derivation of the source clause
interpretation.  This would necessitate the postulation of wild
ambiguity in the source clause, one derivation for each possible case
of subsequent ellipsis.

\subsubsection{Non-subject abstraction}
\label{polarity}

There exist many cases of multiple parallel elements in the source and
target clause; it is very common for ellipsis to involve relations
formed by abstraction of elements other than the interpretation of the
subject noun phrase.

For example, the tense and aspect of the target clause might differ
from that in the source clause:

\enumsentence{
Dan is running for president, and George did last term.}

\noindent The mood can also differ:

\eenumsentence{
\item ``I want to leave.''  ``Well, do.''

\item ``Eat your dinner.''  ``I did.''}

\noindent The two clauses may differ in polarity:

\enumsentence{\label{polarity-ex}
Dan didn't leave, but George did.}

\noindent These examples show that relations of varying arity must be
available as interpretations for elided phrases.  The consequence of
this for theories where the relation available for interpretation of
subsequent ellipsis must be available in the source sentence is,
again, that every sentence which can be the antecedent for subsequent
ellipsis must be many ways ambiguous; an interpretation must be
available for each relation that might be needed to interpret ellipsis
in subsequent discourse.

On the equational analysis, however, such ambiguity is not required.
A single interpretation for the source clause can give rise to any
required interpretation for the target, since there is no inherent
restriction as to the number or nature of the parallel elements
involved in the ellipsis.

Take, for instance, the sentence in (\ref{polarity-ex}).  Here, the
parallel elements are `Dan' and `George' (the subjects), and the
positive and negative polarities, represented semantically as
operators $pos$ and $neg$.\footnote{Other analyses of polarity, for
instance as an implicit argument of the predicate, will yield similar
results.} Thus, the equation

\[P(dan, neg) = \prim{neg}(\fun{left}(\prim{dan}))\]

\noindent can be solved yielding

\[P \mapsto \lamabs{x}{\lamabs{S}{S(\fun{left}(x))}}\]

\noindent and applied to the target parallel elements:

\[P(george, pos) = pos(\fun{left}(george))\eqpunc{.}\]

Of course, gapping and stripping provide abundant examples of
non-subject abstraction involving other arguments or modifiers;
\namecite[page 152]{Reinhart:Anaph} provides this example, which
involves non-subject parallel elements and has both a strict and a
sloppy reading:

\enumsentence{You can keep Rosa in her room for the whole afternoon,
but not Zelda. [Reinhart's (18c)]}

\noindent \namecite[page 275]{Jackendoff:Sem} also discusses examples
involving both subject and non-subject parallelism:

\enumsentence{
Maxwell killed the judge with a silver hammer, and I'd like to do the
same thing to that cop, with a cudgel. [Jackendoff's (6.196)]}

\enumsentence{Fred hung Tessie up in a tree and poured paint on her,
but I bet he wouldn't do it to Sue with glue. [Jackendoff's (6.197)]}

\subsection{Multiple Property Abstraction}

A difficulty in any identity-of-relations analysis, which makes
available only one interpretation for subsequent clauses exhibiting
ellipsis, is seen when a single sentence is the antecedent for the
ellipsis of two different noun phrases.  Consider the following:

\enumsentence{
John finished reading the poem before Bill did, and the short story too.
}

\noindent This sentence has a reading on which John finished reading
both the poem and the short story before Bill finished reading the
poem.  On this reading, the source for both elliptical clauses is the
same clause, `John finished reading the poem.'  To produce a relation
which can be the interpretation for the elided VP whose subject is
Bill, the interpretation for the sentence `John finished reading the
poem' must be derived as:

\[[\lamabs{x}{\mbox{\it finish-reading}\/(x, \mbox{\it the poem})}](john) \]

\noindent so as to make available the property $\lamabs{x}{\mbox{\it
finish-reading}(x, \mbox{\it the poem})}$.  Similarly, an interpretation
must also be produced for the second conjunct `and the short story too'.
On the desired reading, the interpretation for `John finished reading
the poem' must be derived as:

\[[\lamabs{y}{\mbox{\it finish-reading}(john, y)}](\mbox{\it the
poem})\eqpunc{.}\]

\noindent Under an identity-of-relations analysis, the source clause is
deemed ambiguous between the two derivations. They do not
simultaneously exist in a given analysis; only one or the other may be
chosen.  Thus, the reading noted above would not be generable.  On the
other hand, an analysis such as ours allows for the formation of two
different relations from the semantic representation of the first
sentence; the representation of the first sentence does not constrain
the possibilities for construction of such relations.  The next
section provides an example of a similar problem and its derivation in
our framework.

\subsection{Cascaded Ellipsis}

We use the term ``cascaded ellipsis'' to refer to cases of multiple
ellipsis in which one of the elided constituents depends on another
elided constituent for its interpretation.  Analyses dependent on an
identity-of-relations approach generally make available fewer
readings in cascaded ellipsis cases than the analysis presented here;
we believe that the greater number of readings available with our
analysis is in fact warranted.

\namecite{Dahl:SI} provides the
following example (Dahl's (12), an English paraphrase of Scheibe's
(58a) \shortcite{Scheibe:Problem}):

\enumsentence{\label{fool-ex}
John realizes that he is a fool, but Bill does not, even though his
wife does.}

\noindent Dahl claims that this sentence has, among other readings,
the following one:

\enumsentence{John realizes that John is a fool but

Bill does not realize that Bill is a fool, even though

Bill's wife realizes that Bill is a fool
}

\noindent \namecite[page~135 ff.]{Sag:PhD} discusses this example; his claim
is that this reading is not available for this sentence.  We disagree,
and find the reading acceptable.

On our analysis, this reading is
readily available.  Assume that the interpretation for `John realizes
that he is a fool' on the reading under discussion is:

\[ realize(john, \fun{fool}(john)) \]

\noindent This sentence serves as the antecedent for the elided phrase
in the second conjunct, `Bill does not'.  `Bill' and `John' are
parallel elements; for the reading under discussion, second-order
matching solves the equation

\[P(john) = realize(\prim{john},
\fun{fool}(john))\]

\noindent producing, among others, the following property
(corresponding to the sloppy option):

\[P \mapsto \lamabs{x}{realize(x, \fun{fool}(x))} \]

\noindent which is applied
to `Bill'.  The interpretation for the second conjunct as a whole is,
then, the following:\footnote{For simplicity of exposition, we ignore
the fact that the polarities of the two sentences differ.  See
Section~\ref{polarity} for a discussion of this issue.}

\[ realize(bill, \fun{fool}(bill)) \]

\noindent We assume that the second clause may serve as an antecedent
for the elided portion of the third conjunct.  The parallel elements
are `Bill' and `his wife'; the ellipsis equation is\footnote{As the
semantic roles for `Bill' are parasitic on those for `John', we let
the primary occurrences of `Bill' be determined by those of `John' in
the source sentence.}

\[Q(bill) = realize(\prim{bill},
\fun{fool}(bill))\eqpunc{.}\]

\noindent On the reading under discussion, the
strict option is chosen; the property $Q$ applied to the
interpretation of `his wife' is:

\[ \lamabs{x}{realize(x, \fun{fool}(bill))} \]

\noindent The resulting interpretation for the third conjunct is:

\[ realize(\wifeof(bill), \fun{fool}(bill)) \]

Although we have described the derivation of the meaning for this
example in terms of temporal ordering (we resolve the first ellipsis,
using its result to resolve the second), it is important to note that
the analysis is truly order-free.  In essence, we merely set up two
equations in two unknowns and solve them using unification.  The
result, as is typical with declarative, equational methods, does not
depend on solving the equations in a particular order.

Under an identity-of-relations analysis, such as Sag's, the existence
of this reading is problematic, as he notes.  The problem is that
there are conflicting requirements on the form of the semantic
representation of the second clause.

Sag obtains strict and sloppy readings under ellipsis by optionally
applying a rule that replaces the interpretation of a pronoun (which
has an invariant referent and induces a strict reading) by a
lambda-bound variable (inducing a sloppy reading).  The representation
Sag provides for the first two conjuncts is:\footnote{Sag uses the
notation $a, \lamabs{x}{[P]}$ to represent the application of a VP
meaning $\lamabs{x}{[P]}$ to the subject meaning $a$.  The scope of
the negation operator is the entire predication.}

\enumsentence{$John, \lamabs{x}{[\fun{x realize x is a fool}]~but}$

             $\neg Bill, \lamabs{y}{[\fun{y realize y is a fool}]}$}

\noindent Crucially, the interpretation for the pronoun `he' which is
reconstructed in the second conjunct is represented by a bound
variable.

In contrast, for the reading under discussion, the representation for
the second and third conjuncts is:

\enumsentence{$\neg Bill_i, \lamabs{x}{[\fun{x realize
	he$_i$ is a fool}]~\fun{even~though}}$
             $\fun{his$_i$ wife},
	\lamabs{y}{[\fun{y realize he$_i$ is a fool}]}$}

\noindent The strict reading is only available when the option
of replacing the pronoun interpretation with a lambda-bound variable
is not taken.  These conflicting requirements make it impossible for
Sag's analysis---or any identity-of-relations analysis, where the
difference between a strict and a sloppy reading corresponds to a
difference in the form of the semantic representation---to obtain the
reading for this sentence that we (and Dahl) assume exists.

\subsection{Interactions with Quantifier Scope}

As described in Section~\ref{semantic-basics}, quantifier scope is
generated through a mechanism of discharging of assumptions introduced
in the course of a derivation.  The interaction between quantifier
scoping and ellipsis, then, will simply involve the relative
derivational order of discharging such quantifier assumptions and
resolving elliptical constructs.  That is, when we set up the
appropriate instance of the schematic equation $P(s_1, s_2, \ldots,
s_n) = s$, $s$ is the meaning of the source clause, {\it possibly
under one or more undischarged assumptions}.

\subsubsection{Quantification and antecedent-contained ellipsis}
\label{antecedent-contained}

As an example, we consider the two sentences given in
(\ref{quant-egs}).

\eenumsentence{ \label{quant-egs}
\item John greeted every person when Bill did.
\item John greeted every person that Bill did.
}

\noindent As noted by Sag \shortcite{Sag:PhD}, the quantifier scope
possibilities differ for these two sentences: whereas
(\ref{quant-egs}a) admits of two readings, (\ref{quant-egs}b) allows
only one.  Both Sag and Williams \shortcite{Williams:DLF} provide
analyses for these semantic intuitions.

We will take the source of the ellipsis in (\ref{quant-egs}a) to be
the clause `John greeted every person', and the target to be `Bill
did'.  The derivation of interpretations for (\ref{quant-egs}a)
proceeds as follows.  The source clause is interpreted under a
quantifier assumption generated by the subject NP.  (See
Section~\ref{sec:interpretation-of-quantification} for the
derivation.)

\enumsentence{\label{ex:antecedent-contained}
$\langle every\; x\;person(x)\rangle \vdash greet(john, x)$}

\noindent We might discharge the assumption at this point, but we
choose not to in this first scenario.  Consequently, the
interpretation of the full sentence is

\[\langle every\; x\;person(x)\rangle
	\vdash greet(john, x)~when~P(bill)\]

\noindent where $P$ is constrained equationally by virtue of the
interpretation of the source clause:

\[P(john) = greet(\prim{john}, x) \eqpunc{.}\]

\noindent This equation has a single (most general) solution
\[P\mapsto\lamabs{z}{greet(z,x)}\eqpunc{.}\]
\noindent It is this value for $P$ that we will apply to $bill$.
Thus, the interpretation of the full sentence, with ellipsis resolved
is

\[\langle every\; x\;person(x)\rangle
	\vdash greet(john, x)~when~greet(bill, x)\]

\noindent The assumption may now be discharged, yielding the full
interpretation
\[
every(x, person(x), greet(john, x)~when~greet(bill, x))\eqpunc{.}
\]

\noindent This interpretation corresponds to a necessarily
distributive reading, the `individual' reading, in which each person
is simultaneously greeted by John and Bill.

Alternatively, the assumption can be discharged before the ellipsis is
resolved.  Under this scenario, the interpretation of the source clause
is

\[\vdash every(x, person(x), greet(john, x))\eqpunc{.}\]

\noindent The full
sentence, then, is interpreted as

\[\vdash every(x, person(x), greet(john, x))~when~P(bill)\]

\noindent where, again, the interpretation of the source clause is
used to constrain the property $P$:

\[P(john) = every(x,
person(x), greet(\prim{john}, x))\eqpunc{.}\]

\noindent The single value for $P$
is

\[\lamabs{z}{ every(x, person(x), greet(z, x))}\]

\noindent leading to the final interpretation

\[\begin{array}[b]{l}
   every(x, person(x), greet(john, x))~when~\\
   every(x, person(x), greet(bill, x))
  \end{array}\eqpunc{.}\]

\noindent This interpretation yields a `group' reading paraphrasable
as `John greeted every person when Bill greeted every person.'  The
two derivations, then, correspond to just the interpretations noted by
Sag.\footnote{\namecite{Lappin:Ellipsis} proposes a combination of
Cooper storage and copying for interpreting sentences such as
(\ref{quant-egs}a).  His scheme differs substantively from ours in
that he copies both matrix and store from the storage counterpart to
give the interpretation of the elided material.  However, such a
scheme fails to preserve the scope parallelism noted in
Section~\ref{sec:parallelism}.}

Now, we turn to the superficially similar sentence (\ref{quant-egs}b).
We take the source clause to be the entire sentence, and the target to
be, again, `Bill did'.  The interpretation of the sentence before
resolution of ellipsis and discharge of assumptions is

\[\langle every\;x\;person(x) \wedge P(bill) \rangle \vdash
greet(john, x)\eqpunc{.}\]

\noindent (Again, Section~\ref{sec:interpretation-of-quantification}
records a derivation for a similar clause.)

As before, we will resolve the ellipsis by finding solutions for the
equation

\[P(john) = greet(\prim{john}, x)\]

\noindent whose solution assigns
$P$ the relation $\lamabs{z}{greet(z,x)}$.  With this substitution,
the interpretation is

\[\langle every\;x\; person(x) \wedge greet(bill, x)\rangle
\vdash greet(john, x)\]

\noindent which, discharging the assumption, reduces to

\[every(x, person(x) \wedge greet(bill, x), greet(john, x))\eqpunc{.}\]

Alternatively, we might attempt to discharge the assumption before
resolving the ellipsis.  Recall the starting point for the previous
derivation:

\[\langle every\;x\; person(x) \wedge P(bill)\rangle
\vdash greet(john, x)\eqpunc{.}\]

\noindent Discharging the assumption yields

\[
\vdash every(x, person(x) \wedge P(bill), greet(john, x))
\]

\noindent  Resolving the ellipsis, then, involves finding solutions
to the equation

\[P(john) = every(x, person(x) \wedge P(bill),
greet(\prim{john}, x))\eqpunc{.}\]

\noindent Notice that the variable $P$ appears on both sides of this
equation.  For this reason, the derivation runs into problems, for
there simply are no solutions to this equation; no unifier exists for
this pair of typed terms.   Thus, the derivation fails at this
point; the sentence has only the `individual' reading, in agreement
with our judgments.

The computational reflex of the above lack of a solution is a
violation of the so-called ``occurs check'' in the unification
algorithm. This check prevents the construction of unifiers in which
the substitution for a variable contains that variable. In other
words, the occurs check blocks the creation of a circularity (a value
for $P$ containing $P$ itself). It is interesting to note that this is
formally analogous to the syntactic argumentation which Williams
\shortcite{Williams:DLF} uses to eliminate such readings.

In sum, the ellipsis characterization described here allows for
antecedent-contained ellipsis, and predicts correctly the interactions
with quantifier scope.

\subsubsection{Quantification parallelism}
\label{sec:parallelism}

Another implication of the account of quantifier scope given above is
that in cases where the source clause exhibits quantifier scoping
ambiguities, if the quantifiers are separately quantified in the two
clauses (i.e., as in the `group' reading for (\ref{quant-egs}a)) the
scopes in the two clauses must be the same.  For instance, in the
sentence

\enumsentence{
John gave every student a test, and Bill did too.}

\noindent we predict (correctly, we take it) no reading of the form

\[
\begin{array}{l}
every(x, student(x), exists(y, test(y), give(j,x,y))) \\
\wedge exists(y, test(y), every(x, student(x), give(b,x,y)))
\end{array}
\]

\noindent where the two quantifiers take different scopes in the two
clauses.   Consider the possible orderings of ellipsis resolution and
discharging of quantifier assumptions.  If ellipsis resolution occurs
before some of the quantifiers have been discharged, these quantifiers
will scope over {\it both} clauses.  Thus, the only way for both
quantifiers to scope separately is for ellipsis resolution to occur
after both quantifier assumptions are discharged.  But in that case,
the ellipsis equation will include the quantifier order manifest in
the source clause interpretation, and this will be carried over to the
target interpretation.

\subsubsection{Quantification and type raising}
\label{sec:type-raising}

In general, the semantic types of parallel elements must be identical.
In the case of a quantified NP parallel to a non-quantified NP, this
implies that the type of the non-quantified NP must be raised to that
of the quantified type.  As an example, we consider the sentence

\enumsentence{
Every student revised his paper, and then Bill did.}

\noindent This sentence is ambiguous, depending on whether Bill
revises his own paper or each student's paper.

As usual, the ellipsis resolution can occur before or after the
quantifier assumption is discharged.  If the quantifier is discharged
first, we have the meaning given in \ex{1} for the first clause.

\enumsentence{$every(x, student(x), revise(x, \paperof(x)))$}

\noindent To resolve the ellipsis, we need to set up an equation
involving the interpretation of the parallel element in the source,
``every student''. However, for this purpose we cannot directly use
the stored noun phrase interpretation given in \ex{1}.

\enumsentence{$\langle every\;x\;student(x)\rangle\vdash x$}

\noindent This interpretation only makes sense as part of the
derivation of the meaning of the whole source clause. Instead, we
calculate the contribution of ``every student'' to the meaning of the
source clause by examining the effect of applying it to an arbitrary
property $S$.  Combining $S$ with \ex{0}, we obtain first the
interpretation in \ex{1}.

\enumsentence{
$\langle every\;x\;student(x)\rangle\vdash S(x)$}

\noindent The assumption can then be discharged to yield the sentence
meaning in \ex{1}.

\enumsentence{\label{open-quant}$every(x, student(x), S(x))$}

\noindent The contribution of ``every student'' is thus to map an
arbitrary property $S$ to the term in (\ref{open-quant}), that is, the
contribution is the function given in \ex{1}.

\enumsentence{\label{every-student}
$\lamabs{S}{every(x, student(x), S(x))}$}

\noindent It is easy to show that this is equivalent to the usual
generalized quantifier meaning of ``every student''.

To resolve the ellipsis, then, we set up the equation in \ex{1}
involving the meaning~(\ref{every-student}) of the parallel element in
the source, ``every student'':

\enumsentence{
$\begin{array}[t]{l}
P(\lamabs{S}{every(x, student(x), S(x))}) \\
\hspace*{1in} = every(x, student(x), revise(x, \paperof(x)))
\end{array}$ }

\noindent This equation has the solution

\[P \mapsto \lamabs{Q}{Q(\lamabs{x}{revise(x, \paperof(x))})} \eqpunc{.}\]

\noindent The type of the noun phrase meaning
$\lamabs{S}{every(x,~student(x),~S(x))}$ is
$(e~\rightarrow~t)~\rightarrow~t$.  The meaning for ``Bill'' must be
type-raised to $\lamabs{R}{R(b)}$ for type consistency.  The value for
$P$, when applied to the type-raised meaning for ``Bill'', yields the
target meaning

\[revise(b, \paperof(b))\]

\noindent according to which Bill revises his own paper.

On the other hand, if ellipsis resolution occurs first, the derivation
yields

\[\langle every\;x\; student(x)\rangle \vdash revise(x,
\paperof(x))\]

\noindent just before resolution.  The equation

\[P(x) = revise(x, \paperof(x))\]

\noindent admits of a strict interpretation for $P$:

\[P \mapsto \lamabs{y}{revise(y, \paperof(x))}\eqpunc{.}\]

\noindent The meaning for the conjoined sentence before ellipsis
resolution

\[\begin{array}{l}
\langle every\;x\; student(x)\rangle \vdash \\
\hspace*{1in} revise(x, \paperof(x))~and~then~P(b)
\end{array}\]

\noindent reduces, after ellipsis resolution, to

\[\begin{array}{l}
\langle every\;x\; student(x)\rangle \vdash \\
\hspace*{1in} revise(x,\paperof(x))~and~then~revise(b, \paperof(x))
\eqpunc{,}
\end{array} \]

\noindent which, following discharging of the quantifier assumption, becomes

\[every(x, student(x),
\begin{array}[t]{l}
revise(x,\paperof(x))~and~then\\
revise(b, \paperof(x)))
\end{array} \]

\noindent On this reading, Bill revises each student's paper after the
student revises it.  A sloppy reading, on which Bill revises his own
paper, is generable in this way as well; in this particular case,
though, it is logically equivalent to the reading described above, in
which type-lifting of `Bill' is involved.

\subsection{Other Phenomena}

We defer discussion of several other important interactions of
ellipsis and scoping phenomena to a companion paper in preparation.
In that paper we intend to discuss, in addition to a more detailed
explication of the mainstream quantifier cases:

\begin{description}
\item[Scope ambiguities with indefinites:]  Indefinites give rise to
several readings under ellipsis, depending on their scope and the
relative order of ellipsis resolution and discharge of the indefinite.
This sentence, for example, has three readings:

\begin{quote}
John lost a book he owned, and so did Bill.
\end{quote}

On the first reading, John and Bill lost the same book; on the second
reading, John and Bill each lost one of John's books, possibly
distinct; and on the third reading Bill lost one of his own books.

\item[De dicto/de re ambiguities:]  Similar ambiguities are found in
sentences with opaque verbs.  This sentence has three readings:

\begin{quote}
Bill wants to read a good book and John does too.
\end{quote}

Where the first conjunct has a de dicto reading, the second conjunct
does also; where the first conjunct has a de re reading, however,
there are two readings for the sentence as a whole, depending on
whether or not Bill and John want to read the same book.

\item[`Canadian flag' examples:]  \namecite{Hirshbuhler:Ellipsis}
discusses examples such as the following:

\begin{quote}
A Canadian flag was hanging in front of each window, and an American
one was too.
\end{quote}

Examples such as these illustrate that the subject of the source
clause need not take widest scope in VP ellipsis.  On our analysis,
these examples, like those in Section~\ref{sec:type-raising}, involve
matching of higher than second order.

\end{description}

\section{A Comparison of Approaches}

In order to more fully explicate the differences between our approach
to ellipsis resolution and other approaches, we analyze a single
example in detail from the perspective of several previous proposals.
Rather than make the comparison in the respective notations of the
original proposals, we normalize those notations by using lambda
terms uniformly.  When the analyses are viewed in this way, several
apparently different analyses are seen to generate the same set of
readings in the same manner, despite their having originally been
stated in differing notations.

We will use the following example, discussed at length by Gawron and
Peters \shortcite{GP:Anaph}:

\enumsentence{\label{GP-ex}
John revised his paper before the teacher did, and Bill did too.}

\noindent We follow Gawron and Peters in directing attention to the
reading of the first conjunct where `John' and `his' corefer, and of
the second elliptical clause where its source is the entire previous
sentence.  The following six readings exhaust the set of readings
generated by any of the analyses (including our own) that we discuss.
We present them with paraphrases for reference.

\eenumsentence{\label{revise-ex}
\item
{\samepage
\begin{tabbing}
$\before($\=$revise(john,~\paperof(john)),$ \\
         \>$revise(teacher,~\paperof(teacher)))~and $ \\
$\before($\>$revise(bill,~\paperof(bill)),$ \\
         \>$revise(teacher,~\paperof(teacher))) $
\end{tabbing}

Each person revised his own paper.}

\item
{\samepage
\begin{tabbing}
$\before($\=$revise(john,~\paperof(john)),$ \\
         \>$revise(teacher,~\paperof(john)))~and $ \\
$\before($\>$revise(bill,~\paperof(john)),$ \\
         \>$revise(teacher,~\paperof(john))) $
\end{tabbing}

Each person revised John's paper.}

\item
{\samepage
\begin{tabbing}
$\before($\=$revise(john,~\paperof(john)),$ \\
         \>$revise(teacher,~\paperof(john)))~and $ \\
$\before($\>$revise(bill,~\paperof(bill)),$ \\
         \>$revise(teacher,~\paperof(bill))) $
\end{tabbing}

John and then the teacher revised John's paper; Bill and then
the teacher revised Bill's paper. }

\item
{\samepage
\begin{tabbing}
$\before($\=$revise(john,~\paperof(john)),$ \\
         \>$revise(teacher,~\paperof(teacher)))~and $ \\
$\before($\>$revise(bill,~\paperof(john)),$ \\
         \>$revise(teacher,~\paperof(teacher))) $
\end{tabbing}

John and Bill both revised John's paper before the teacher revised the
teacher's paper.}

\item
{\samepage
\begin{tabbing}
$\before($\=$revise(john,~\paperof(john)),$ \\
         \>$revise(teacher,~\paperof(john)))~and $ \\
$\before($\>$revise(bill,~\paperof(bill)),$ \\
         \>$revise(teacher,~\paperof(john))) $
\end{tabbing}

John and Bill revised their own papers before the teacher
revised John's paper.}

\item
{\samepage
\begin{tabbing}
$\before($\=$revise(john,~\paperof(john)),$ \\
         \>$revise(teacher,~\paperof(john)))~and $ \\
$\before($\>$revise(bill,~\paperof(john)),$ \\
         \>$revise(teacher,~\paperof(bill))) $
\end{tabbing}

John and then the teacher revised John's paper; Bill revised John's
paper before the teacher revised Bill's paper.  }}

\noindent As we will see, the equational analysis that we propose in
this paper is the most profligate of the analyses, potentially
generating all six of these readings, though restrictions might
eliminate certain of these.

\subsection{Zero-Reading Analyses}

The strictest version of an identity-of-relations analysis
requires that a lambda term used in the derivation of the meaning of
the source clause be used in the derivation of the target clause
meaning (either by copying or deletion under identity).  Under such
an analysis, the pertinent level of semantic representation of the
source clause to use in the target clause derivation is that before
beta reduction has occurred, as beta reduction eliminates the
function-typed lambda terms.  For the sentence `John revised his
paper', the unreduced meaning representation is one of:

\[\lamabs{x}{revise(x, \paperof(john))}(john)\]

\noindent or

\[\lamabs{x}{revise(x, \paperof(x))}(john)\eqpunc{.}\]

\noindent corresponding to the strict and sloppy readings,
respectively.  In forming the meaning of the first target clause
`before the teacher did', we use whichever term $P$ is made
available by the first clause, generating $P(teacher)$.  An issue
remains as to how the two clause meanings are then combined to form a
single sentence meaning.  The most natural method, direct
coordination, would yield (for the sloppy reading):\footnote{Here and
elsewhere, we uniformly rename bound variables apart for clarity.}

\[\before(\hspace{-1ex}\begin{array}[t]{l}
	\lamabs{x}{revise(x, \paperof(x))}(john),\\
	\lamabs{y}{revise(y, \paperof(y))}(teacher))
	 \end{array}\]

\noindent corresponding to a syntactic analysis under which the
adverbial clause is attached at the S level.  However, this is not
itself of the form appropriate for being the source of a later
ellipsis, that is, a function applied to the subject meaning.  Thus,
under this analysis, the second ellipsis, `and Bill did too', would be
uninterpretable.  (Recall that we are ignoring the readings in which
the source for the second ellipsis is merely `John revised his paper'.
Such a reading would be possible, although it would be strict or
sloppy dependent on the interpretation  of the other two
clauses.)

\subsection{Two-Reading Analyses}

The second ellipsis is not, of course, uninterpretable, so we attempt
to design a meaning representation for its source that is of the
appropriate form.  A first method is to place the meaning of the
`before' clause within the meaning of its source VP:

\[[\lamabs{x}{\before(\hspace{-1ex}\begin{array}[t]{l}
			revise(x, \paperof(x)),\\
			\lamabs{y}{revise(y, \paperof(y))}(teacher))] (john)
				\end{array}}\]

\noindent (We will discuss a second method in
Section~\ref{3-readings}.)  The exact mechanism for constructing such
a reading is not specified here.  It may seem especially problematic,
as the source and target terms are not alphabetic variants in this
notation.  Nonetheless, if this were the meaning representation of the
source of the second ellipsis, it would allow for a sloppy reading of
the target; this would correspond to the sloppy
reading~(\ref{revise-ex}a).  The strict variant

\[[\lamabs{x}{\before(\hspace{-1ex}\begin{array}[t]{l}
			revise(x, \paperof(john)),\\
			\lamabs{y}{revise(y, \paperof(john))}(teacher))] (john)
				\end{array}}\]

\noindent would yield reading~(\ref{revise-ex}b).

An alternative method of subordinating the `before' clause meaning
maintains the alphabetic variance property of the two clauses.  If we
assume, counterintuitively, that the {\it before} operator takes a
property and a truth value to a property (that is, it is of type
$((e~\rightarrow~t)~\times~t)~\rightarrow~(e~\rightarrow~t)$), we can
form the meaning representation

\[[\before(\hspace{-1ex}\begin{array}[t]{l}
			\lamabs{x}{revise(x, \paperof(x))},\\{}
  			[\lamabs{y}{revise(y, \paperof(y))}](teacher))](john)
			\end{array}
\]

\noindent and similarly for the strict case.  Again, the mechanism is
a bit mysterious (though less so) and readings~(\ref{revise-ex}a)
and~(\ref{revise-ex}b) would be generated for the sentence as a whole.

The analyses of \namecite{Sag:PhD} and \namecite{Williams:DLF}, although
they do not consider cases such as these, might be reasonably viewed
as generating these readings in much this way.  Similarly, the
analysis presented by \namecite{Roberts:PhD} and phrased in terms of
DRT generates these two readings (as discussed by Gawron and Peters).

\subsection{Three-Reading Analyses}
\label{3-readings}

The representation for the source of the second ellipsis might be
extrapolated along different lines.  In particular, the interpretation
for each verb phrase, including the compound one, might be given by an
overt abstraction.  This would correspond to a syntactic analysis
under which the adverbial clause is adjoined to the main clause verb
phrase, with one verb phrase meaning appearing as a subconstituent of
the other.  For the sloppy reading~(\ref{revise-ex}a), we would have

\[\lamabs{u}{[\before(\hspace{-1ex}\begin{array}[t]{l}
	\lamabs{x}{revise(x, \paperof(x))}(u),\\
	\lamabs{z}{revise(z, \paperof(z))}(teacher)](john)\eqpunc{,}
				\end{array}}
\]

\noindent and for the strict (\ref{revise-ex}b)

\[\lamabs{u}{[\before(\hspace{-1ex}\begin{array}[t]{l}
	\lamabs{x}{revise(x, \paperof(john))}(u),\\
	\lamabs{z}{revise(z, \paperof(john))}(teacher)](john)\eqpunc{.}
				\end{array}}
\]

\noindent This representation has the benefit of being uniform,
preserving alphabetic variance, and assigning a more attractive type
to {\it before}.  It also allows for a third reading when used as the
source for the second elliptical clause.  The second argument to {\it
revise} might be---in addition to {\it j} (the strict option) or {\it
x} (the locally sloppy option)---the reabstracted subject meaning {\it
u}.  This leads to a globally sloppy meaning~(\ref{revise-ex}c).

\[\lamabs{u}{[\before(\hspace{-1ex}\begin{array}[t]{l}
	\lamabs{x}{revise(x, \paperof(u))}(u),\\
	\lamabs{z}{revise(z, \paperof(u))}(teacher)](john)
				\end{array}}
\]

This analysis, which generates three readings for the example
sentence, is essentially the analysis developed by Gawron and Peters
within a situation-theoretic framework. (The final reading corresponds
to the second argument of {\it revise} being ``absorbed'' by the
lambda operator.)  Our transliteration makes clear that, for this case
at least, situation-semantics machinery is not necessary to yield the
readings in question; an extrapolation of Sag's or Williams's analyses
might achieve the same result.  Of course, other aspects of the Gawron
and Peters analysis depend intrinsically on the situation-theoretic
foundation.

\subsection{Six-Reading Analyses}

The three readings provided by the Gawron and Peters analysis seem to
exhaust the possibilities for an identity-of-relations approach.
Our analysis produces six readings for the example sentence.

Let us examine in detail how one of the readings for this sentence,
(\ref{revise-ex}c) above, is obtained under the equational analysis of
ellipsis.  Assume that the semantics for `John revised his paper' is:

\[revise(john, \paperof(john)) \]

\noindent
The first conjoined sentence then will have the meaning

\[\before(revise(john,\paperof(john)), P(teacher))\]

\noindent
under the constraint

\[P(john) = revise(\prim{john}, \paperof(john)) \eqpunc{.}\]

\noindent
The second elliptical clause takes its source to be the whole first
conjunction.  Thus, its interpretation will be

\[\before(revise(john, \paperof(john)), P(teacher)) \wedge
Q(bill)\]

\noindent under the constraint

\[Q(john) = \before(revise(\prim{john}, \paperof(john)),
P(teacher))\eqpunc{.}\]

These two equations in two unknowns ($P$ and $Q$) are solved, as
usual, by higher-order unification; we will take the equation for $P$
first.  One solution is to take the strict reading for $P$,

\[P =
\lamabs{x}{revise(x, \paperof(john))}\]

\noindent leading to the following interpretation for the second equation:

\[Q(john) =
\before(\hspace{-1ex}\begin{array}[t]{l}revise(\prim{john}, \paperof(john)),\\
revise(teacher, \paperof(john))) \end{array}
\]

\noindent
This equation, in turn, has a solution

\[Q =
\lamabs{x}{ \before(\hspace{-1ex}\begin{array}[t]{l}
	revise(x,~\paperof(x)), \\
        revise(teacher,~\paperof(x)))
		\end{array}}
\]

\noindent The semantics for this reading of the sentence as a whole is:

\[
\begin{array}{l}
\before(\hspace{-1ex}\begin{array}[t]{l}revise(john,~\paperof(john)), \\
                          revise(teacher,~\paperof(john)))~and
       \end{array}\\
\before(\hspace{-1ex}\begin{array}[t]{l}revise(bill,~\paperof(bill)), \\
                          revise(teacher,~\paperof(bill)))
       \end{array}
\end{array}\]

Our analysis allows for readings that are missing under the analyses
discussed above because it is not an identity-of-relations analysis;
interpretation of ellipsis does not involve copying the interpretation
of a constituent in the source.

\subsection{Five-Reading Analyses}

In Section~\ref{sec:ant-anaphor-constraints}, we discuss a restriction
on unifiers that uniformly eliminates certain readings of elliptical
clauses.  This restriction, when applied to the example at hand,
eliminates reading~(\ref{revise-ex}f).  Thus, our analysis, strictly
speaking, generates only five of the six combinatorially possible
readings of the sentence.

\subsection{Four-Reading Analyses}

An unpublished analysis attributed to Hans Kamp (personal
communication to Mark Gawron and Stanley Peters, cited by Gawron and
Peters \shortcite{GP:Anaph}) and couched in DRT assigns four readings
to the sentence, and does so by eliminating the identity-of-relations
assumption.  In Kamp's analysis, as in our own, ambiguities between
strict and sloppy readings do not arise from ambiguity in the source
clause; the source has only a single interpretation.  Essentially,
Kamp makes a copy of the discourse representation structure of the
source, and then imposes constraints identifying the participants in
the source and target copies. These constraints must be applied in a
symmetric manner.  If a sloppy interpretation constraint applies to
one copied discourse entity, it must apply to all; similarly for a
strict interpretation constraint.  Gawron and Peters mention a
possible extension to Kamp's analysis that allows for the generation
of all six of the readings listed above by relaxing the symmetry
requirement.  We refer the reader to the discussion by Gawron and
Peters for a fuller description of Kamp's proposal.

Insofar as Kamp's analysis can be fleshed out, his analysis and ours
make the same predictions as to the class of readings available in
cases of cascaded ellipsis.  Readings missing under other analyses are
available for our analysis and his.  The particular syntactic
operations that Kamp (under Gawron and Peters's reconstruction)
presupposes have no particular foundation other than efficacy.  Our
analysis can be seen as providing an argument for the operational view
implicit in Kamp's analysis based on the underlying equational
characterization of elliptical constructions.  This equational
foundation, as we have seen, articulates with other semantic phenomena
in ways not appreciated in the previous research.

\subsection{Summary}

In summary, each analysis differs in the number of analyses that are
predicted for the given sentence.  Here is a scorecard.

\begin{center}
\begin{tabular}{|c|c|}
\hline
{\it Method} & {\it \# readings} \\
\hline
Sag, Roberts & 2 \\
Gawron and Peters & 3 \\
Kamp & 4 (or 6) \\
equational & 5 \\
\hline
\end{tabular}
\end{center}

The single reading that our method derives that remains underived by
others is that in (\ref{revise-ex}e).  Clearly, this reading is
difficult, if not impossible, to dig out (although plausibility
considerations play a large role here).  However, we have seen
examples that demonstrate that the reason for its absence in the other
analyses is faulty.  For instance, its elimination on the basis of an
identity-of-relations analysis, as Roberts or Gawron and Peters would
have it, has repeatedly been seen to be too strong.

Rather, the pertinent distinction in differentiating the first four readings
from the last two is that the resolution of the second elliptical
construction in the last two readings must treat the parallel
structures that ellipsis applies to in a non-parallel fashion.  We
conjecture that such non-parallel cases are highly dispreferred, if
not disallowed entirely.

In order to test this hypothesis, we can try to construct an example
which pragmatically favors this dispreferred reading to see whether it
is obtainable.  The following sentence is parallel in structure to
sentence~(\ref{GP-ex}):
\enumsentence{
Dewey announced his victory after the newspapers did, but so did Truman.}

\noindent A reasonable, historically accurate reading
for this sentence may be represented as:

\enumsentence{$
\begin{array}[t]{l}
after(\hspace{-1ex}\begin{array}[t]{l}announce(Dewey,~Dewey$'$s~victory), \\
                         announce(newspaper,~Dewey$'$s~victory))~but
     \end{array} \\
after(\hspace{-1ex}\begin{array}[t]{l}announce(Truman,~Truman$'$s~victory), \\
                         announce(newspaper,~Dewey$'$s~victory))
     \end{array}
\end{array} $ }

\noindent
That is, Dewey claimed victory for himself after the newspapers
announced Dewey's victory, but Truman also claimed victory after Dewey
was announced the winner by the newspapers.  This reading is parallel
to reading~(\ref{revise-ex}e) described above.

Opinions differ as to the acceptability of this reading.  One's
opinion in this case can be seen as a litmus determining whether
parallelism of the sort violated here is required, or merely
preferred.

\section{Problematic Cases}

The following issues are problematic for most analyses of ellipsis
interpretation.  We present them, along with our conjectured
solutions, to codify the range of phenomena that analyses of ellipsis
might account for and to provide a preliminary guess as to their
possible solutions in our framework.

\subsection{Non-Syntactic Parallelism}
\label{sec:sem-parallel}

Our analysis of ellipsis resolution presupposes identification of the
source of the ellipsis and the parallel structuring of the source and
target.  This division of labor between identification of parallelism
and resolution of ellipsis is purposeful, as the factors involved in
the solution of the two problems are quite different.  Although
determining the parallelism may seem to be a purely syntactic
operation, much like the matching that goes on at the semantic level,
this similarity is illusory.  Cases of semantic or pragmatic
parallelism also exist.  These cases are particularly
problematic for theories of ellipsis in which the interpretation of an
elided phrase is presumed to correspond to the interpretation of some
syntactic constituent in the source clause, as is the case in most
identity-of-relations analyses.

\subsubsection{Semantic parallelism}

Examples of ellipsis exist in which the parallelism is between the
``logical subject'' (i.e., passive agent) in the source clause and the
surface subject in the target clause:

\eenumsentence{
\item  A lot of this material can be presented in a fairly informal
and accessible fashion, and often I do.  \cite[page 41]{Chomsky:GenEnt}

\item It should be noted, as Dummett does, $\ldots$ (example due to Ivan Sag)}

\noindent Similar examples involving ``so-anaphora'' are also
found:

\eenumsentence{
\item  The formalisms are thus more aptly referred to as information- or
constraint-based rather than unification-based, and we will do so
here. \cite[page 2]{Shieber:PhD}

\item  It is possible that this result can be derived from some
independent principle, but I know of no theory that does so.
\cite[page 664]{Mohanan:FuncAnaphControl}}

\noindent  Examples of this type are ubiquitous, but seem
to be confined pragmatically to cases where the source clause states a
general fact or rule, and the target clause provides a specific
instance of this fact or rule.

 Examples of passive/active parallelism are not confined
to those in which the source and target appear in the main clause.  In
the following example (due to \namecite{Wescoat:StrictSloppy}),
parallelism of the heads of the main clause subjects forces a
parallelism of arguments of the modifying relative clauses; {\it
John}, the object of the relative clause in the source sentence, is
parallel to {\it Bill}, the subject of the relative clause in the
target sentence:

\enumsentence{
The policeman who arrested John failed to read him his rights, and
so, for that matter, did the one Bill got collared by.}

Our analysis does not require that the property provided as the
interpretation for the elided portion of the target clause in examples
like those above correspond to the interpretation of any constituent
in the source clause.  It is not clear that there is any analysis
available for examples of this sort within a theory in which the
interpretation for elided phrases must be that of some constituent in
the source clause, as is the case in most identity-of-relations
analyses.

 Other cases of semantic/thematic parallelism can also
arise; sentence \ex{1} is from instructions on a bottle of Agree
shampoo:

\enumsentence{Avoid getting shampoo in eyes---if it does, flush
thoroughly with water.}

\noindent Syntactically, the parallel elements are the object of the
source clause and the subject of the target; thematically, these
elements are the ``theme'' arguments of the intransitive/causative
{\it get} verb pair.

Other combinations of logical-subject/surface-subject parallelism do
not seem to arise.  The following examples, where the parallel
elements are intended to be the surface subject in the source clause
and the logical subject in the target, are ungrammatical:

\eenumsentence{
\item  $\ast$Dummett notes, as it should be, $\ldots$

\item  $\ast$We will refer to the formalisms as information- or
constraint-based, as they more aptly are.
}

\noindent However, the following sentence (due to Peter Sells) has a
similar structure, yet seems to be more acceptable:

\enumsentence{
John completed the assignment faster than it ever had been in the
history of the school.}

\noindent To the extent that this sentence is grammatical, it
illustrates that either the source or target clause can contain a
logical subject which is parallel to a surface subject in the other
clause.  Although examples such as these are often
restricted in their distribution, they demonstrate that the
parallelism between elements in the source and target clause need not
be confined to surface syntactic parallelism.

\subsubsection{Pragmatic parallelism}

More extreme cases exist in which arbitrary information may need to be
brought to bear to determine the appropriate parallelism between
source and target.  \namecite{Webber:PhD} cites some such
cases:

\eenumsentence{
\label{ex:webber-sentences}
\item Irv and Mary want to dance together but Mary can't since her husband
is here.

\item Mary wants to go to Spain and Fred wants to go to Peru, but because
of limited resources, only one of them can.
}

\noindent \noindent  Recovery of the pertinent properties in these
sentences requires nonlinguistic knowledge concerning social norms and
economic processes.  In example~(\ref{ex:webber-sentences}b), for
instance, the implicit property that only one of Mary and Fred can
have is {\it to do what he or she wants to do}.  Other attested
examples include:\footnote{We are indebted to James McCawley and
Bonnie Webber for bringing these examples to our attention.}

\enumsentence{Fortunately, the first person to die in 1990 and the
first couple to file for divorce in 1990 were allowed to do so
anonymously.  \cite{roeper}}

\enumsentence{Amid applause at the Congress of the   Russian
Federation (RSFSR), Mr. Yeltsin put forward a bill setting Russian law
above the law of the Soviet Union -- something Mr. Gorbachev, as
Soviet president, declared unconstitutional when Estonia, Latvia and
Lithuania did it last year.  \cite{guardian}}

Sentences of this sort illustrate that, to a greater or lesser extent,
relations involved in the resolution of anaphoric processes such as
ellipsis can be made available contextually.  Identity-of-relations
analyses allow for only the simplest cases of resolution of elided
constituents, since the only mechanism that is available to provide an
interpretation for the target is that of copying an interpretation
from the source.  Our approach goes beyond identity-of-relations
analyses by allowing for the construction of new relations on the
basis of old ones; the use of unification to construct relations is,
as we have seen, more powerful and more flexible than copying.

Even for cases of what \namecite{SagHankamer:DeepSurf} call ``surface
anaphora'', such as verb phrase ellipsis, resolution is possible to
relations that are pragmatically determined or influenced, as the
examples in (\ref{ex:webber-sentences}) show.  To interpret these
examples, not only the apparatus we introduce for constructing new
relations but also pragmatic knowledge must be brought to bear.

\subsection{Further Constraints on Relation Formation}

Cases in which sloppy but not strict readings are available might seem
to be problematic for an analysis like the one presented here.  Below
we will examine cases of control and reflexivization in which
exclusively sloppy readings are available.  Solutions will be proposed
which do not involve constraining the process by which relations are
formed as interpretations for elided constituents.

However, there are other cases in which readings are unexpectedly
unavailable; these cases generally involve multiple occurrences of
pronouns whose antecedent is a parallel element in the source clause.
It seems that a constraint is necessary on the possibilities for
forming relations in cases such as these.

\subsubsection{Obligatory sloppy readings}

\paragraph{Control}

In general, only sloppy readings are available for sentences involving
control.  The following sentence is not ambiguous:

\enumsentence{
John tried to run, and Bill did too.}

\noindent There is no reading according to which Bill tries to bring
it about that John runs.

Chierchia \shortcite{Chierchia:Control,Chierchia:Inf}, noting facts of
this type, proposes that the semantic type of a controlled verb phrase
is a property rather than a proposition.  That is, for a sentence like
``John tried to run'', the correct semantic representation would be
(a) and not (b):

\eenumsentence{
\item[a.] $try(john,\lamabs{x}{run(x)})$

\item[b.] $try(john,run(john))$
}

\noindent If Chierchia's hypothesis is correct, the lack of a strict
reading is predicted; in the representation in (a), there is only a
single occurrence of ``$john$'', and a strict reading is impossible to
produce.

However, there are reasons to doubt the adequacy of an analysis like
this one.  First, anaphors whose antecedent is the subject of a
controlled VP can give rise to both a sloppy and a strict reading
under ellipsis.  Consider this sentence:

\enumsentence{
\label{eg:property}
John tried to kill himself before Bill did.
}

\noindent Since the reflexive ``himself'' cannot be bound to a higher
clause subject, its antecedent must be the subject of ``kill''.  On the
property analysis, the interpretation for the first conjunct would
then be:

\enumsentence{\label{property-anal-ex}
$try(john,\lamabs{x}{kill(x,x)})$
}

However, we find this sentence to have two readings, corresponding to
the following paraphrases:\footnote{Additional readings arise if the
source clause for the ellipsis is taken to be not the matrix ``John
tried to kill himself'', but the VP complement ``to kill himself'':
\enumsentence[(a)]{
John tried to kill himself before Bill killed himself.}
\enumsentence[(b)]{
John tried to kill himself before Bill killed John.}

Similar comments apply to these examples.}

\eenumsentence{
\item[a.] John tried to kill himself before Bill tried to kill himself.

\item[b.] John tried to kill himself before Bill tried to kill John.
}

\noindent The second of these readings would not be obtainable
given~(\ref{property-anal-ex}) as the source clause meaning.

\namecite{Zec:Control} demonstrates another problem for Chierchia's
property analysis.  The data she cites bear on Chierchia's claim that
there is a correlation between the syntactic form and the semantic
type of complements: that complements of type VP are properties, while
complements of type S$'$ are propositions.  Chierchia predicts that
there would be no case in which an S$'$ complement gives rise to only
a sloppy reading.

Zec discusses cases of obligatory control in Serbo-Croatian, showing
that there are cases of obligatory control into verbal complements
that are of the syntactic category S$'$ rather than VP.  For example,
the verb {\it poku{\v{s}}ati} ``try'' takes an S$'$ complement, and
yet only a sloppy reading is possible in the following sentence
\cite[page 143]{Zec:Control}:

\enumsentence{\label{serbo-sloppy-eg}
\gloss{Petar}{Petar}
\gloss{je}{Aux}
\gloss{poku{\v{s}}ao}{tried}
\gloss{da}{that}
\gloss{postane}{become}
\gloss{predsednik}{president}
\gloss{a}{and}
\gloss{to}{it}
\gloss{je}{Aux}
\gloss{poku{\v{s}}ala}{tried}
\gloss{i}{too}
\gloss{Marija}{Marija}
\hfilbreak
``Petar tried to become president and Marija tried it too.''
}

\noindent This sentence means that Marija tried to bring it about that
Marija (not Petar) become president.

There seem to be two possibilities with regard to the Serbo-Croatian
data.  The first is to deny that there is a necessary correlation
between the syntactic type of the complement and its semantic type.
The claim would be in that case that, although the syntactic type of
the complement of ``try'' is an S$'$ in Serbo-Croatian, semantically
it is a property.  In that case, the lack of a strict reading would be
predicted for the Serbo-Croatian case, just as it is for the English
case; the problem illustrated by example~(\ref{eg:property}) would
remain, however.

Another option, and the one that Zec takes, is to posit an obligatory
coreference relation between the subject of ``try'' and the subject of
its complement clause; this relation would presumably be induced by
the control verb.  If this option is taken, it would presumably force
the abstraction of both arguments at the same time under second-order
matching.

In our terms, the obligatory coreference relation and the obligatory
sloppy readings Zec discusses would mean that controlled occurrences
are primary in both Serbo-Croatian and English.  Thus,
in~(\ref{eg:property}), the interpretation would be

\[try(\prim{john},
kill(\prim{john}, john))\eqpunc{,}\]

\noindent which, when taken to be a source for ellipsis, would
generate only two solutions to the equation

\[P(john) = try(\prim{john}, kill(\prim{john}, john))\eqpunc{,}\]

\noindent manifesting a sloppy reading for the controlled subject
occurrence and either a strict or a sloppy reading for the reflexive
occurrence, as required.

\paragraph{Reflexivization}

\namecite{SZZ:Anaph} present a number of cases of reflexivization
in which only sloppy readings are available.  The Dutch reflexive {\it
zich } is such a case \cite[page~182]{SZZ:Anaph}:

\enumsentence{
\gloss{Zij}{She}
\gloss{verdedigde}{defended}
\gloss{zich}{herself}
\gloss{beter}{better}
\gloss{dan}{than}
\gloss{Peter}{Peter}
\hfilbreak
``She defended herself better than Peter.''
}

\noindent Sells et al.\ characterize reflexive constructions involving
only sloppy readings as ``closed predicate'' constructions.  They
discuss only examples in which the reflexive appears in object
position, with its antecedent being the subject of the same clause.

We might assume, then, that for the examples they discuss, the
presence of the reflexive correlates with the operation of a semantic
relation-reducing rule, one which semantically ``intransitivizes'' the
verb.  In the Dutch case, then, the presence of {\it zich} signals a
change in the meaning of the verb from the meaning in (a) to the one
in (b):

\eenumsentence{
\item[a.] $\lamabs{x}{\lamabs{y}{\fun{defend}}(x, y)}$

\item[b.] $\lamabs{x}{\fun{defend-self}}(x)$
}

\noindent A solution of this type would work for all cases of
obligatory sloppy readings for reflexives that are described by Sells
et al., since they consider only cases where the reflexive and its
antecedent are arguments of the same predicate, in which a
relation-reducing operation can apply.

However, this solution would be inappropriate in cases where the
reflexive and its antecedent are clearly arguments of different
predicates, where a relation-reducing operation cannot apply.
Although such cases are difficult to find, it may be that the
Serbo-Croatian reflexive {\it sebe} (genitive {\it svoje}) is such a
case.

The following sentence has only a sloppy reading (Draga Zec, personal
communication):

\enumsentence{
\gloss{Petar}{Petar}
\gloss{je}{Aux}
\gloss{sakrio}{hid}
\gloss{sto}{one}
\gloss{hiljada}{hundred}
\gloss{dolara}{dollars}
\gloss{ispod}{underneath}
\gloss{svoje}{self's}
\gloss{ku{\'{c}}e}{house}
\gloss{a}{and}
\gloss{to}{that}
\gloss{je}{Aux}
\gloss{u{\v{c}}inio}{did}
\hfilbreak
\gloss{i}{also}
\gloss{Pavle}{Pavle}
\hfilbreak
``Petar hid one hundred dollars underneath self's house, and Paul did
(that) too.''}

\noindent The only reading available for this sentence is that Paul
hid one hundred dollars under his own house.

We postulate that the reflexive {\it sebe} in Serbo-Croatian engenders
primary as opposed to secondary occurrences, which would then be
subject to the primary occurrence constraint. This would also be true
of the English reflexive for those speakers who find that strict
readings with reflexives are unacceptable.

In short, a variety of syntactic constructions give rise to multiple
primary occurrences of parallel elements: control, both of the type
seen in English and of the type seen in Serbo-Croatian, and
reflexivization in some dialects of English and in Serbo-Croatian.

\subsubsection{Antecedent-anaphor constraints}
\label{sec:ant-anaphor-constraints}

\paragraph{Missing readings with multiple occurrences of anaphora}

In cases where there are two pronouns coreferent with the
parallel element in the source, one might expect that each pronoun
would give rise to either a strict or a sloppy reading, giving a
total of four readings for the target clause.  This does not seem to
be the case, however; one of the readings is systematically missing.

\namecite{Dahl:HOS} notes that the following sentence, with two
occurrences of pronouns in the source clause, has only three and not
four interpretations:

\eenumsentence{\label{bill-harry-eg}
\item[a.] Bill believed that he loved his wife, and Harry did too.

\item[b.] {\it Harry believed that Bill loved Bill's wife.}

\item[b$''$.] {\it Harry believed that Harry loved Harry's wife.}

\item[b$''$.] {\it Harry believed that Harry loved Bill's wife.}

\item[b$'''$.] {\it $\ast$ Harry believed that Bill loved Harry's wife.}

}

\noindent
In this case, the missing reading corresponds to the following unifier
for~(\ref{missing-reading-eq}a):

\eenumsentence{\label{missing-reading-eq}
\item{$P(bill) = believe(\prim{bill},love(bill,~\wifeof(bill)))$}

\item{$P \mapsto \lamabs{x}{believe(x, love(bill,~\wifeof(x)))} $}
}

Other examples illustrate a similar phenomenon.
\namecite[page~183]{Sag:PhD} observes that this sentence has only three
readings, not four:

\eenumsentence{\label{edith-eg}
\item[a.] Edith said that finding her husband nude had upset
her, and Martha did too.

\item[b.] {\it Martha said that finding Martha's husband nude had upset
Martha.}

\item[b$'$.] {\it Martha said that finding Edith's husband nude had upset
Edith.}

\item[b$''$.] {\it Martha said that finding Edith's husband nude had upset
Martha.}

\item[b$'''$.] {\it $\ast$ Martha said that finding Martha's husband nude
had upset Edith.}}

\noindent The interpretation paraphrased in \ex{0}b$'''$ is missing.
The unifier for P for the missing reading is:

\eenumsentence{
\item{$P(edith) =
say(\prim{edith},~upset(\findingnude(\husbandof(edith)),edith))$}

\item{$P \mapsto
\lamabs{x}{say(x,~upset(\findingnude(\husbandof(x)),edith))} $}
}

Examples~(\ref{bill-harry-eg}) and~(\ref{edith-eg}) illustrate a
constraint on relation formation.  In example~(\ref{bill-harry-eg}),
the position corresponding to the pronoun ``his'' may not be
abstracted unless the position corresponding to ``he'' is also
abstracted.  The constraint does not seem to correlate with linear
order of the pronouns, though; in example~(\ref{bill-harry-eg}) the
position corresponding to the rightmost pronoun may not be abstracted
unless the position corresponding to the leftmost pronoun is also
abstracted.  The reverse is true in example~(\ref{edith-eg}), however,
where it is the position corresponding to the rightmost pronoun that
must be extracted.

Although these examples show that the proper generalization about the
ordering between pronominal positions does not have to do with linear
order, they are consistent with the hypothesis that the ordering
correlates with depth of syntactic embedding.  In both
example~(\ref{bill-harry-eg}) and example~(\ref{edith-eg}), if the
position corresponding to a more deeply embedded pronoun is abstracted
over, the position corresponding to a less deeply embedded pronoun
must also be abstracted over.

However, example~(\ref{revise-eg-again}) shows that the ordering
between positions may not be dependent on syntactic facts at all.
Recall that sentence~(\ref{GP-ex}), repeated here, was not associated
with the reading in~(\ref{revise-eg-again}b), paraphrased
in~(\ref{revise-eg-again}c):

\eenumsentence{\label{revise-eg-again}
\item John revised his paper before the teacher did, and Bill did too.

\item
{\samepage
\begin{tabbing}
$\before($\=$revise(john,~\paperof(john)),$ \\
         \>$revise(teacher,~\paperof(john)))~and $ \\
$\before($\>$revise(bill,~\paperof(john)),$ \\
         \>$revise(teacher,~\paperof(bill))) $
\end{tabbing}}

\item John and then the teacher revised John's paper; Bill revised John's
paper before the teacher revised Bill's paper.}

\noindent On the excluded reading, the source clause is ``John revised
his paper before the teacher did'', and the target clause is ``Bill
did too''.  ``John'' and ``Bill'' are the parallel elements.  The
unifier for P for this reading is:

\eenumsentence{
\item{\begin{tabbing}
$P(john) =
\before(\hspace{-1ex}\begin{array}[t]{l}revise(\prim{john},~\paperof(john)),\\
                                        revise(teacher,~\paperof(john)))
                     \end{array}$
\end{tabbing}}

\item{
$P \mapsto
\lamabs{x}{\before(\hspace{-1ex}\begin{array}[t]{l}revise(x,~\paperof(john)),\\
                                                   revise(teacher,\paperof(x)))
                                \end{array}}$
}}

These various missing readings can be captured by positing a linking
relationship between the semantics of pronouns and that of their
antecedents, and generalizing it to include the relation between the
semantics of terms induced by ellipsis and that of their source
parallel element.  Under a suitable definition of this generalized
antecedent linking, all of the cases here can be captured by requiring
that if an occurrence is abstracted over, so must its generalized
antecedent.

\paragraph{Apparent Syntactic Constraints}

Finally, we turn to some simple examples that seem to lack any
readings whatsoever.  Consider the following examples, where ``Mary''
is taken to be the antecedent of ``she'':\footnote{Examples of this
type are due to \namecite{FM:Ellipsis}.}

\eenumsentence{
\item $\ast$ John gave Mary everything she did.

\item $\ast$ John likes Mary, and she does too.}

\noindent These judgments would follow from an analysis on which
syntactic structure is copied from the source to the target, as the
sentences with copies in place violate constraints on binding.

\eenumsentence{
\item $\ast$ John gave Mary everything she$_i$ gave Mary$_i$.

\item $\ast$ John likes Mary, and she$_i$ likes Mary$_i$ too.}

However, a simple copying analysis faces problems in accounting for
grammatical examples of similar structure:

\eenumsentence{
\item John got to Sue$_i$'s apartment before she$_i$ did.

\item John voted for Sue$_i$ because she$_i$ told him to.
}

On a copying analysis, these examples would be incorrectly predicted
to be ungrammatical, just as their copied versions are:

\eenumsentence{
\item *John got to Sue$_i$'s apartment before she$_i$ got to Sue$_i$'s
apartment.

\item *John voted for Sue$_i$ because she$_i$ told him to vote for Sue$_i$.
}

\noindent The resolution of this puzzle remains an open
question, as does its incorporation in the present analysis; for
discussion of the problem, see \namecite{Hellan:Anaph},
\namecite{FM:Ellipsis}, and \namecite{Kitagawa:Ellipsis}.

Another example that seems to argue for a quite superficial analysis
of ellipsis is the following, due to Yoshihisa Kitagawa (personal
communication to Peter Sells):

\enumsentence{John thinks that Mary will revise his paper before Bill
will.}

\noindent on the reading in which Mary revises John's paper and Bill
revises his own paper.  We find the intuition questionable, but it is
clearly problematic for our (and many others') analysis if the reading
is deemed to be available.

Finally, examples which seem to argue for the presence of a gap in the
ellipsis site include the following:

\enumsentence{
*John met everyone that Peter wondered when he could.
\cite[p.~511]{Haik:Ellipsis}}

\enumsentence{
*Tom visited everyone who told Sue where to. \cite[p.~218]{Haik:PhD}}

\noindent On the assumption that long-distance dependencies are
syntactically constrained and that subjacency violations involve an
improper syntactic relation between a filler and a gap, these examples
indicate that the ellipsis site contains a gap at syntactic structure.

\section{Conclusion}

The underlying idea in the analysis of ellipsis that we have presented
here---namely, the construction of higher-order equations on the basis
of parallel structures, and their solution by unification---has been
exemplified primarily by the verb-phrase ellipsis construction.
However, many other elliptical phenomena and related phenomena subject
to multiple readings akin to the strict and sloppy readings discussed
here may be analyzed using the same techniques.  The ambiguities in
cleft sentences such as

\enumsentence{It is Dan who loves his wife.}

\noindent and interpretation of ``only'' with respect to its focus, as in

\enumsentence{Only Dan loves his wife.}

\noindent as well as more standard elliptical phenomena such as
stripping and comparative deletion can be analyzed in this way as
well, making a broad range of predictions as to the space of possible
readings and their interaction with other semantic phenomena.  It
remains for future work to test these potential applications more
fully.

We adduce three advantages of the analysis of elliptical constructions
presented here over previous alternatives.  First, it is in certain
respects simpler, in that it requires no postulation of otherwise
unmotivated ambiguities in the source clause.  Second, it is more
accurate in its predictions, especially in allowing readings
disallowed in identity-of-relations analyses.  Third, it is
methodologically preferable in that the analysis follows directly from
a semantic statement of the ellipsis problem with little stipulation.
The operation on which it relies, higher-order unification, is
semantically sound in that the results it produces are determined by
the meanings of phrases directly rather than by the form of the
representations encoding those meanings, as operations of deletion or
copying of portions of such representations are.

\section*{Acknowledgements}

We would like to thank the following people for helpful discussion:
Hiyan Alshawi, Sam Bayer, Joan Bresnan, Mark Gawron, Kris Halvorsen,
Dan Hardt, Julia Hirschberg, David Israel, Mark Johnson, Ron Kaplan,
Lauri Karttunen, Shalom Lappin, Richard Larson, Peter Ludlow, John
Maxwell, Richard Oehrle, Stanley Peters, Hub Prust, Steve Pulman, Mats
Rooth, Ivan Sag, Peter Sells, Gregory Ward, Michael Wescoat, Annie
Zaenen, Draga Zec, and two anonymous reviewers.  The written comments
alone that we received prior to publication ran to well over 50 pages,
longer, in fact, than the paper itself.  We regret that we could not
include discussion of all the important issues that they raised.


\makeatletter
\def\thesourcebibliography#1{\section*{Sources of Attested Examples\@mkboth
 {Sources of Attested Examples}{Sources of Attested Examples}}\list
 {}{\setlength{\labelwidth}{0pt}\setlength{\leftmargin}{\parindent}
 \setlength{\itemindent}{-\parindent}}
 \def\newblock{\hskip .11em plus .33em minus -.07em}
 \sloppy\clubpenalty4000\widowpenalty4000
 \sfcode`\.=1000\relax}
\let\endthesourcebibliography=\endlist
\makeatother

\newpage

\bibliographystyle{fullname}

\begin{thebibliography}{}

\bibitem[\protect\citename{van Benthem}1989]{Vanbenthem:Categorial}
Johan~van Benthem.
\newblock 1989.
\newblock Categorial grammar and type theory.
\newblock {\em Journal of Philosophical Logic}.
\newblock To appear.

\bibitem[\protect\citename{Chierchia}1983]{Chierchia:Control}
Gennaro Chierchia.
\newblock 1983.
\newblock Outline of a semantic theory of (obligatory) control.
\newblock In Michael Barlow, Daniel Flickinger, and Michael Wescoat, editors,
  {\em Proceedings of the West Coast Conference on Formal Linguistics 2}, pages
  19--31. Stanford Linguistics Association, Stanford University.

\bibitem[\protect\citename{Chierchia}1984]{Chierchia:Inf}
Gennaro Chierchia.
\newblock 1984.
\newblock Anaphoric properties of infinitives and gerunds.
\newblock In Mark Cobler, Susannah Mac{K}aye, and Michael Wescoat, editors,
  {\em Proceedings of the West Coast Conference on Formal Linguistics 3}, pages
  28--39. Stanford Linguistics Association, Stanford University.

\bibitem[\protect\citename{Chierchia}1988]{Chierchia:Dynamic}
Gennaro Chierchia.
\newblock 1988.
\newblock Dynamic generalized quantifiers and donkey anaphora.
\newblock In M.~Krifka, editor, {\em Proceedings of the 1988 T{\"{u}}bingen
  Conference}. Seminar f{\"{u}}r Nat{\"{u}}rliche-Sprachiche Systeme der
  Universit{\"{a}}t T{\"{u}}bingen, November.

\bibitem[\protect\citename{Cooper}1983]{Cooper:Quant}
Robin Cooper.
\newblock 1983.
\newblock {\em Quantification and Syntactic Theory}, volume~21 of {\em Synthese
  Language Library}.
\newblock D. Reidel, Dordrecht.

\bibitem[\protect\citename{Dahl}1972]{Dahl:SI}
{\"O}sten Dahl.
\newblock 1972.
\newblock On so-called `sloppy identity'.
\newblock In {\em Gothenburg Papers in Theoretical Linguistics}, volume~11.
  University of G{\"{o}}teborg.

\bibitem[\protect\citename{Dahl}1974]{Dahl:HOS}
{\"O}sten Dahl.
\newblock 1974.
\newblock How to open a sentence: Abstraction in natural language.
\newblock In {\em Logical Grammar Reports, No. 12}. University of
  G{\"{o}}teborg.

\bibitem[\protect\citename{Fiengo and May}1990]{FM:Ellipsis}
Robert Fiengo and Robert May.
\newblock 1990.
\newblock Anaphora and ellipsis.
\newblock {\sc ms}, City University of New York and University of California,
  Irvine.

\bibitem[\protect\citename{Friedman}1975]{Friedman:Equality}
Harvey Friedman.
\newblock 1975.
\newblock Equality between functionals.
\newblock In R.~Parikh, editor, {\em Lecture Notes in Mathematics 453}, pages
  22--37. Springer-Verlag, Berlin, Germany.

\bibitem[\protect\citename{Gawron and Peters}1990]{GP:Anaph}
Mark Gawron and Stanley Peters.
\newblock 1990.
\newblock {\em Anaphora and quantification in {S}ituation {S}emantics}.
\newblock CSLI/University of Chicago Press, Stanford University.
\newblock CSLI Lecture Notes, Number 19.

\bibitem[\protect\citename{Gazdar \bgroup et al.\egroup }1985]{GKPS:GPSG}
Gerald Gazdar, Ewan Klein, Geoffrey~K. Pullum, and Ivan~A. Sag.
\newblock 1985.
\newblock {\em Generalized Phrase Structure Grammar}.
\newblock Harvard University Press, Cambridge, MA.

\bibitem[\protect\citename{Groenendijk and
  Stockhof}1987]{GroenendijkStockhof:Dynamic}
J.~Groenendijk and M.~Stockhof.
\newblock 1987.
\newblock Dynamic {M}ontague {G}rammar.
\newblock Paper presented at the Workshop on Discourse Representation Theory,
  Stuttgart, West Germany, December.

\bibitem[\protect\citename{Ha{\"{\i}}k}1985]{Haik:PhD}
Isabelle Ha{\"{\i}}k.
\newblock 1985.
\newblock {\em The Syntax of Operators}.
\newblock {Ph.D.} thesis, MIT.

\bibitem[\protect\citename{Ha{\"{\i}}k}1987]{Haik:Ellipsis}
Isabelle Ha{\"{\i}}k.
\newblock 1987.
\newblock Bound {VP}s that need to be.
\newblock {\em Linguistics and Philosophy}, 10:503--530.

\bibitem[\protect\citename{Hankamer and Sag}1976]{SagHankamer:DeepSurf}
Jorge Hankamer and Ivan~A. Sag.
\newblock 1976.
\newblock Deep and surface anaphora.
\newblock {\em Linguistic Inquiry}, 7(3):391--428.

\bibitem[\protect\citename{Heim}1982]{Heim:PhD}
Irene Heim.
\newblock 1982.
\newblock {\em The Semantics of Definite and Indefinite Noun Phrases}.
\newblock {Ph.D.} thesis, University of Massachusetts-Amherst.

\bibitem[\protect\citename{Hellan}1988]{Hellan:Anaph}
Lars Hellan.
\newblock 1988.
\newblock {\em Anaphora in {N}orwegian and the Theory of Grammar}.
\newblock Foris Publications, Dordrecht.

\bibitem[\protect\citename{Hindley and Seldin}1986]{HindleySeldin:Lambda}
J.~Roger Hindley and Jonathon~P. Seldin.
\newblock 1986.
\newblock {\em Introduction to Combinators and $\lambda$-Calculus}.
\newblock Cambridge University Press, Cambridge, England.

\bibitem[\protect\citename{Hirschberg and
  Ward}1991]{HirschbergWard:StrictSloppy}
Julia Hirschberg and Gregory Ward.
\newblock 1991.
\newblock Accent and bound anaphora.
\newblock {\em Cognitive Linguistics}.
\newblock To appear.

\bibitem[\protect\citename{Hirshb{\"{u}}hler}1982]{Hirshbuhler:Ellipsis}
Paul Hirshb{\"{u}}hler.
\newblock 1982.
\newblock {VP} deletion and across-the-board quantifier scope.
\newblock In James Pustejovsky and Peter Sells, editors, {\em Proceedings of
  NELS 12}. GLSA, University of Massachusetts-Amherst.

\bibitem[\protect\citename{Hobbs and Shieber}1987]{HS:Quant}
Jerry~R. Hobbs and Stuart~M. Shieber.
\newblock 1987.
\newblock An algorithm for generating quantifier scopings.
\newblock {\em Computational Linguistics}, 13:47--63.

\bibitem[\protect\citename{Huet and Lang}1978]{HuetLang:SecondOrder}
G\'{e}rard Huet and Bernard Lang.
\newblock 1978.
\newblock Proving and applying program transformations expressed with
  second-order patterns.
\newblock {\em Acta Informatica}, 11(1):31--55.

\bibitem[\protect\citename{Huet}1975]{Huet:HOU}
G\'{e}rard Huet.
\newblock 1975.
\newblock A unification algorithm for typed $\overline{\lambda}$-calculus.
\newblock {\em Theoretical Computer Science}, 1:27--57.

\bibitem[\protect\citename{Jackendoff}1972]{Jackendoff:Sem}
Ray~S. Jackendoff.
\newblock 1972.
\newblock {\em Semantic Interpretation in Generative Grammar}.
\newblock MIT Press, Cambridge, MA.

\bibitem[\protect\citename{Kamp}1981]{Kamp:TSR}
Hans Kamp.
\newblock 1981.
\newblock A theory of truth and semantic representation.
\newblock In Jeroen Groenendijk, Theo Janssen, and Martin Stokhof, editors,
  {\em Formal Methods in the Study of Language}, pages 277--321, Amsterdam.
  Mathematical Centre.

\bibitem[\protect\citename{Kitagawa}1991]{Kitagawa:Ellipsis}
Yoshihisa Kitagawa.
\newblock 1991.
\newblock Copying identity.
\newblock {\em Natural Language and Linguistic Theory}.
\newblock To appear.

\bibitem[\protect\citename{Lambek}1980]{Lambek80}
Joachim Lambek.
\newblock 1980.
\newblock From $\lambda$-calculus to cartesian closed categories.
\newblock In J.P. Seldin and J.R. Hindley, editors, {\em To H.B. Curry: Essays
  on Combinatory Logic, Lambda Calculus and Formalism}, pages 375--402.
  Academic Press, London.

\bibitem[\protect\citename{Lappin}1984]{Lappin:Ellipsis}
Shalom Lappin.
\newblock 1984.
\newblock {VP} anaphora, quantifier scope, and logical form.
\newblock {\em Linguistic Analysis}, 13(4):273--315.

\bibitem[\protect\citename{Pereira}1990]{Pereira:SemComp}
Fernando C.~N. Pereira.
\newblock 1990.
\newblock Categorial semantics and scoping.
\newblock {\em Computational Linguistics}, 16(1):1--10.

\bibitem[\protect\citename{Pulman}1988]{Pulman:CLE}
S.~G. Pulman.
\newblock 1988.
\newblock A contextual reasoning and cooperative response framework for the
  {C}ore {L}anguage {E}ngine.
\newblock Internal report, SRI Cambridge, Cambridge, England.

\bibitem[\protect\citename{Reinhart}1983]{Reinhart:Anaph}
Tanya Reinhart.
\newblock 1983.
\newblock {\em Anaphora and Semantic Interpretation}.
\newblock University of Chicago Press, Chicago.

\bibitem[\protect\citename{Roberts}1987]{Roberts:PhD}
Craige Roberts.
\newblock 1987.
\newblock {\em Modal Subordination, Anaphora, and Distributivity}.
\newblock {Ph.D.} thesis, University of Massachusetts-Amherst.

\bibitem[\protect\citename{Sag}1976]{Sag:PhD}
Ivan~A. Sag.
\newblock 1976.
\newblock {\em Deletion and Logical Form}.
\newblock {Ph.D.} thesis, MIT.

\bibitem[\protect\citename{Scheibe}1973]{Scheibe:Problem}
Traugott Scheibe.
\newblock 1973.
\newblock Zum {P}roblem der grammatisch relevanten {I}dentit{\"{a}}t.
\newblock In F.~Kiefer and N.~Ruwet, editors, {\em Generative Grammar in
  {E}urope}, pages 482--527. D.~Reidel Publishing Company, Dordrecht.

\bibitem[\protect\citename{Schubert and Pelletier}1982]{SchubertPelletier82}
Len Schubert and F~J. Pelletier.
\newblock 1982.
\newblock From {E}nglish to logic: Context-free computation of `conventional'
  logical translations.
\newblock {\em American Journal of Computational Linguistics}, 10:165--176.
\newblock Reprinted in Grosz {\it et al.}, 1986.

\bibitem[\protect\citename{Sells \bgroup et al.\egroup }1987]{SZZ:Anaph}
Peter Sells, Annie Zaenen, and Draga Zec.
\newblock 1987.
\newblock Reflexivization variation: Relations between syntax, semantics, and
  lexical structure.
\newblock In Masayo Iida, Stephen Wechsler, and Draga Zec, editors, {\em
  Working Papers in Grammatical Theory and Discourse Structure}, pages
  169--238. CSLI/University of Chicago Press, Stanford University.
\newblock CSLI Lecture Notes, Number 11.

\bibitem[\protect\citename{Steedman}1990]{Steedman:Gapping}
Mark~J. Steedman.
\newblock 1990.
\newblock Gapping as constituent coordination.
\newblock {\em Linguistics and Philosophy}, 13(2):207--263.

\bibitem[\protect\citename{Webber}1978]{Webber:PhD}
Bonnie~Lynn Webber.
\newblock 1978.
\newblock {\em A Formal Approach to Discourse Anaphora}.
\newblock {Ph.D.} thesis, Harvard University.

\bibitem[\protect\citename{Wescoat}1989]{Wescoat:StrictSloppy}
Michael Wescoat.
\newblock 1989.
\newblock Sloppy readings with embedded antecedents.
\newblock {\sc ms}, Stanford University.

\bibitem[\protect\citename{Williams}1977]{Williams:DLF}
Edwin Williams.
\newblock 1977.
\newblock Discourse and logical form.
\newblock {\em Linguistic Inquiry}, 8(1):101--139.

\bibitem[\protect\citename{Zec}1987]{Zec:Control}
Draga Zec.
\newblock 1987.
\newblock On obligatory control in clausal complements.
\newblock In Masayo Iida, Stephen Wechsler, and Draga Zec, editors, {\em
  Working Papers in Grammatical Theory and Discourse Structure}, pages
  139--168. CSLI/University of Chicago Press, Stanford University.
\newblock CSLI Lecture Notes, Number 11.

\end{thebibliography}

\newpage

\begin{thesourcebibliography}{}

\bibitem[\protect\citename{Chomsky}1982]{Chomsky:GenEnt}
Noam Chomsky.
\newblock 1982.
\newblock {\em Noam Chomsky on the Generative Enterprise}.
\newblock Foris Publications, Dordrecht.

\bibitem[\protect\citename{Mohanan}1983]{Mohanan:FuncAnaphControl}
K.~P. Mohanan.
\newblock 1983.
\newblock Functional and anaphoric control.
\newblock {\em Linguistic Inquiry}, 14(4):641--674.

\bibitem[\protect\citename{Rettie}1990]{guardian}
John Rettie.
\newblock 1990.
\newblock Yeltsin wants {Russia} to go it alone.
\newblock {\em The Guardian}, 23 May.

\bibitem[\protect\citename{Roeper}1990]{roeper}
R.~Roeper.
\newblock 1990.
\newblock {\it Chicago Sun-Times}, 8 January.
\newblock Cited by James McCawley, ``1990 Linguistic Flea Circus'', unpublished
  manuscript.

\bibitem[\protect\citename{Shieber}1989]{Shieber:PhD}
Stuart~M. Shieber.
\newblock 1989.
\newblock {\em Parsing and Type Inference for Natural and Computer Languages}.
\newblock {Ph.D.} thesis, Stanford University.

\end{thesourcebibliography}

\end{document}